%% file: main.tex
\documentclass{article}

\usepackage[preprint]{neurips_2026}

\usepackage[utf8]{inputenc} %
\usepackage[T1]{fontenc} %
\usepackage{hyperref} %
\usepackage{url} %
\usepackage{booktabs} %
\usepackage{amsfonts} %
\usepackage{nicefrac} %
\usepackage{microtype} %

\usepackage{amsmath}
\usepackage{graphicx}
\usepackage{multirow}
\usepackage{bm}
\usepackage[table]{xcolor}
\usepackage{xspace}
\usepackage{enumitem}
\usepackage{subcaption}
\usepackage{dashrule}
\usepackage{algorithm}
\usepackage{algorithmicx}
\usepackage{algpseudocode}
\usepackage{cleveref}

\title{FASTER: Rethinking Real-Time Flow VLAs}

\author{%
Yuxiang Lu$^{1,2}$ \quad Zhe Liu$^{1,*}$ \quad Xianzhe Fan$^{1}$ \quad Zhenya Yang$^{1}$ \quad Jinghua Hou$^{1}$ \\
\textbf{Junyi Li}$^{1}$ \quad \textbf{Kaixin Ding}$^{1}$ \quad \textbf{Hengshuang Zhao}$^{1,\dagger}$ 
\\
$^{1}$The University of Hong Kong \quad $^{2}$ACE Robotics\\
$^{*}$Project Leader \quad $^{\dagger}$Corresponding Author \\
\texttt{https://innovator-zero.github.io/FASTER} \\
}

\newcommand{\tctrl}{\Delta t_{\text{ctrl}}}
\newcommand{\tinfer}{\Delta t_{\text{infer}}}
\newcommand{\texec}{\Delta t_{\text{exec}}}
\newcommand{\treact}{\Delta t_{\text{react}}}
\newcommand{\dreact}{D_\text{react}}
\newcommand{\A}{\mathbf{A}}
\newcommand{\Z}{\mathbf{Z}}
\newcommand{\method}{FASTER\xspace}
\newcommand{\eg}{\textit{e.g.}\xspace}
\newcommand{\ie}{\textit{i.e.}\xspace}
\newcommand{\vs}{\textit{vs.}\xspace}

\algrenewcommand\algorithmicrequire{\textbf{Input:}}
\algrenewcommand\algorithmicensure{\textbf{Output:}}

\definecolor{mygreen}{HTML}{d5e8d4}

\begin{document}
  \maketitle

  \begin{figure}[h]
    \vspace{-10mm}
    \centering
    \includegraphics[width=0.95\linewidth]{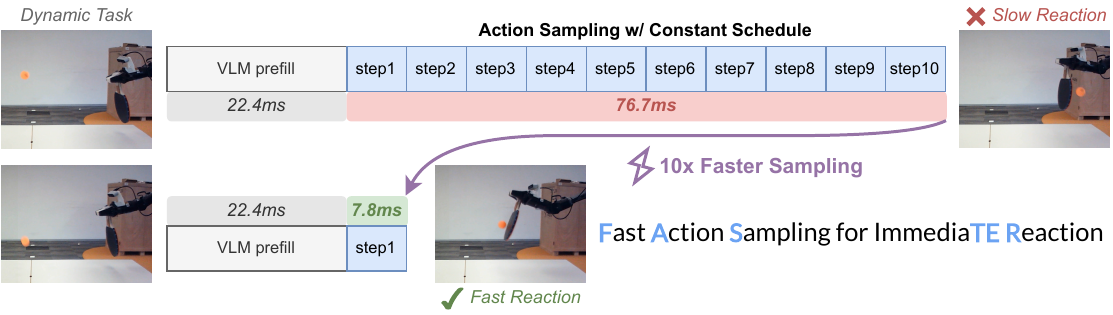}
    \vspace{-2mm}
    \caption{We propose \method to alleviate the reaction latency bottleneck in
    action chunking flow policies. By compressing the sampling iterations of the
    immediate reaction into a single step, \method (bottom) achieves a \textbf{10$\times$
    acceleration} compared to original $\pi_{0.5}$ and X-VLA (top). This enables
    real-time responsiveness in highly dynamic real-world tasks such as playing
    table tennis. Moreover, \method is a plug-and-play solution for flow-based VLAs that integrates seamlessly into the standard fine-tuning pipeline and requires no architectural modifications.}
    \label{fig:teaser}
  \end{figure}

  \begin{abstract}
    Real-time execution is crucial for deploying Vision-Language-Action (VLA) models
    in the physical world. Existing asynchronous inference methods primarily
    optimize trajectory smoothness, but neglect the critical latency in reacting
    to environmental changes. By rethinking the notion of reaction in action
    chunking policies, this paper presents a systematic analysis of the factors governing
    reaction time. We show that reaction time follows a uniform distribution
    determined jointly by the Time to First Action (TTFA) and the execution
    horizon. Moreover, we reveal that the standard practice of applying a
    constant schedule in flow-based VLAs can be inefficient and forces the
    system to complete all sampling steps before any movement can start, forming
    the bottleneck in reaction latency. To overcome this issue, we propose \textbf{F}ast
    \textbf{A}ction \textbf{S}ampling for Immedia\textbf{TE} \textbf{R}eaction (\textbf{FASTER}).
    By introducing a Horizon-Aware Schedule, FASTER adaptively prioritizes near-term
    actions during flow sampling, compressing the denoising of the immediate reaction
    by tenfold (\eg, in $\pi_{0.5}$ and X-VLA) into a single step, while preserving
    the quality of long-horizon trajectory. Coupled with a streaming client-server
    pipeline, FASTER substantially reduces the effective reaction latency on real
    robots, especially when deployed on consumer-grade GPUs. Real-world experiments,
    including a highly dynamic table tennis task, prove that FASTER unlocks substantially improved
    real-time responsiveness for generalist policies, enabling rapid generation
    of accurate and smooth trajectories.
  \end{abstract}

  \section{Introduction}
  \label{sec:intro}

  The paradigm of robot learning is undergoing a profound transformation with the
  advent of Vision-Language-Action (VLA) models~\cite{zhang2025pure, shao2025large,
  ma2024survey, sapkota2025vision}. By formulating continuous motor control as a
  generative sequence modeling problem, recent approaches leveraging diffusion models~\cite{ddpm,
  ddim} and flow matching~\cite{flowmatching} for action chunking have achieved unprecedented
  capabilities in dexterous robotic manipulation tasks~\cite{pi06, gr00t, xvla, lingbotvla}.
  As research focus shifts from simulation to real-world physical deployment, real-time
  capability has become increasingly paramount.

  Existing real-time execution methods primarily address the ``stop-and-wait'' issue
  in standard synchronous inference~\cite{dp} for action chunking policies~\cite{smolvla}.
  By introducing an asynchronous pipeline, the robot can initiate the next inference
  request before the current action chunk is exhausted, thereby eliminating
  inter-chunk pauses and enhancing motion continuity~\cite{rtc}. While state-of-the-art
  advances in asynchronous inference strategy~\cite{trainingrtc, remac, vlash, legato,
  vla-rail} further reinforce trajectory smoothness, these methods largely
  overlook another essential dimension of real-time embodied intelligence: \textbf{reaction}.
  Beyond smooth execution, a practical VLA system must promptly and precisely
  respond to dynamically changing physical environments. Delayed reactions to unexpected
  perturbations create a perilous ``blind spot'' in closed-loop control,
  limiting the effectiveness of generalist policies in open-world scenarios.

  Our in-depth analysis of the inference pipeline in \Cref{sec:analysis} reveals
  that reaction time is not a trivial constant determined by inference latency. Instead,
  it should be modeled as a random variable following a uniform distribution, due
  to the stochastic timing of external events relative to robot controller. We
  further illustrate that existing asynchronous methods are inherently limited, and
  a collaborative enhancement in both perception-execution latency and the frequency
  of inference-execution cycle is entailed to acquire truly responsive behavior.

  We then revisit a common design in flow-based VLAs: a constant timestep
  schedule across the action chunk, which allocates an equal number of sampling steps
  to every action. Under this scheme, the full multi-step denoising process must
  be completed before any action can be dispatched, severely inflating the
  reaction delay. Considering the intrinsic causal structure of physical interaction,
  near-term actions are more tightly coupled with current observations and
  typically lie in a significantly narrower solution space. Our pilot study in \Cref{sec:pilot_study}
  supports this intuition. We clearly observe that early actions follow
  straighter interpolation paths and attain precise estimation of clean actions
  within only a few sampling steps, whereas a constant schedule over-samples these
  dimensions. This naturally raises a key question: \textit{since earlier
  actions are easier to predict than later ones, can flow-based VLAs generate these
  latency-critical actions with fewer sampling steps for immediate reaction?}

  To address these challenges, we propose \textbf{F}ast \textbf{A}ction \textbf{S}ampling
  for Immedia\textbf{TE} \textbf{R}eaction (\textbf{\method}), a simple yet
  effective method applicable to flow-based VLAs~\cite{pi0, xvla} without
  architectural modifications or additional training cost. As shown in \Cref{fig:teaser},
  \method aims to accelerate the sampling process of leading actions, as quantified
  by the newly introduced Time to First Action (TTFA) metric for reactivity.
  Concretely, a Horizon-Aware Schedule (HAS) is incorporated to decouple the local
  denoising timestep for each frame within the chunk. HAS adaptively allocates
  more aggressive sampling steps to near-term actions while maintaining a slower
  schedule for long-horizon ones. Consequently, the model can output the
  immediate action as fast as one-step sampling, while largely preserving long-term trajectory
  accuracy.

  Beyond algorithmic acceleration, \method also catalyzes a paradigm shift from
  conventional asynchronous pipeline to a streaming client-server interaction, wherein
  early actions can be dispatched to the robot controller instantly upon completion.
  While the robot executes these initial movements, the VLA model continues refining
  subsequent actions in parallel and progressively replenishes the client's action
  buffer. Real-world evaluations on two GPU platforms~(\textit{i.e.}, RTX 4060 and
  RTX 4090) demonstrate that \method substantially reduces inference latency, as
  reflected by lower TTFA, while simultaneously boosting inference-execution
  cycle frequency through the synergization of streaming output and early-stopping
  strategies. Real-robot experiments further confirm the superior reaction capability
  of \method, even when deployed on resource-constrained GPUs, offering a general
  and promising path toward genuinely real-time VLAs.

  Our contributions are summarized as: (1) We present a systematic analysis of
  reaction attributes in action chunking VLA policies, revealing the inherent limitations
  of existing methods for real-time responsiveness. (2) We propose the \method
  framework, capitalizing on a Horizon-Aware Schedule that prioritizes immediate actions
  during flow matching sampling, effectively compressing TTFA to one-step
  sampling without sacrificing prediction quality. (3) We design a streaming client-server
  interface with early stopping, jointly trimming the delay and accelerating the
  closed loop of inference-execution. (4) Extensive experiments on real robots demonstrate
  significant improvements in reaction capability and promising performance in dexterous
  action generation for manipulation tasks.

  \section{Analysis on Action Chunking Policy Inference}
  \label{sec:analysis}

  \begin{figure}[t]
    \centering
    \includegraphics[width=0.95\linewidth]{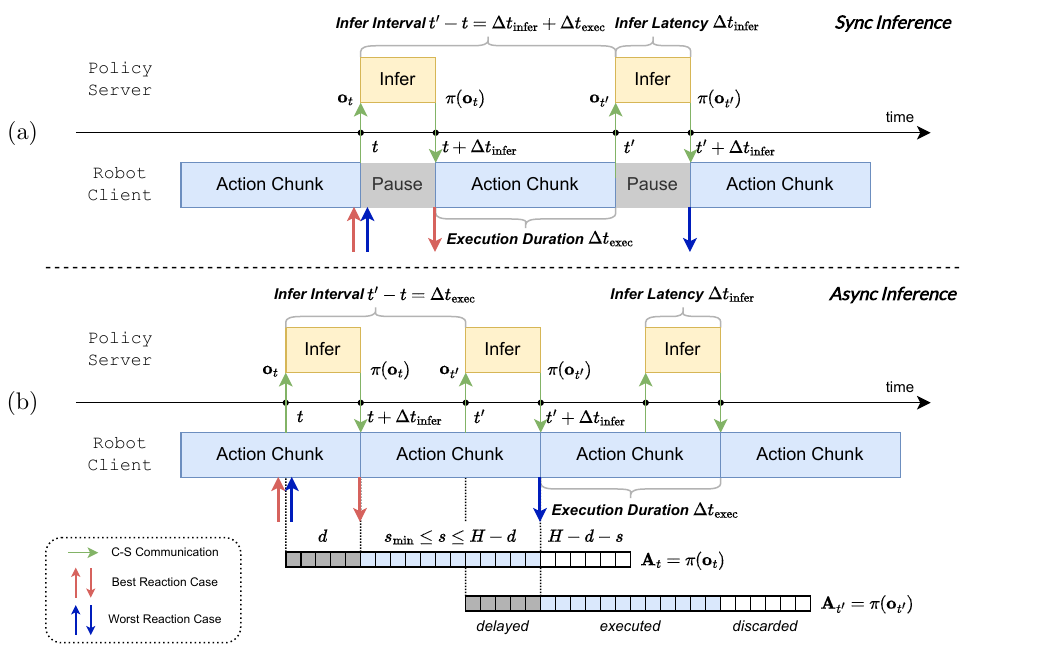}
    \caption{Temporal pipelines of (a) synchronous and (b) asynchronous
    inference in a robotic system composed of an action chunking policy server
    and a robot client. As indicated by the best and worst cases, reaction time
    depends on both inference latency and the interval between consecutive
    inference-execution cycles. We also illustrate the decomposition of two adjacent
    action chunks to clarify the discretized inference delay $d$ and the execution
    horizon $s$ (bounded by $s_{\text{min}}$ and $H-d$) in the asynchronous client.}
    \label{fig:infer_study}
    \vspace{-3mm}
  \end{figure}

  Action chunking is a standard method in VLA policies~\cite{dp, act, pi0}.
  Given a policy $\pi$, the model processes an observation $\mathbf{o}_{t}$ at real-world
  time $t$ to predict a sequence of future actions $\A_{t}= [\bm{a}_{t}, \bm{a}_{t+1}
  , \dots , \bm{a}_{t+H-1}]$, where $H$ denotes the \textit{prediction horizon} specified
  by the policy. In practice, instead of executing the entire action chunk, it
  is common to execute only $s$ actions, then trigger a new inference and discard
  the remaining actions. $s$ is referred to as the \textit{execution horizon}~\cite{rtc}.

  Deploying a VLA policy on a physical robot typically utilizes a client-server architecture,
  consisting of a policy server for model inference and a robot client for motor
  control. After initialization, the server remains active to process incoming
  requests from the client and returns predictions with a certain latency, giving
  rise to two interaction paradigms: synchronous and asynchronous inference.

  We first define several time quantities to assist analysis of the pipeline:
  \begin{itemize}[noitemsep, topsep=0pt, leftmargin=*]
    \item \textit{Control period} $\tctrl:= 1/f$. In robotic systems, the
      controller operates at a specific frequency $f$ (\eg, 30Hz), corresponding
      to a fixed period between consecutive operations, such as executing action
      or triggering inference.

    \item \textit{Inference latency} $\tinfer$. This is defined as the time interval
      between transmitting an observation and receiving the predicted actions on
      the client side. It encompasses model inference, network communication,
      pre- and post-processing, memory I/O, and other system overheads. For
      analytical convenience, we model the total latency as a constant.
      Following prior work~\cite{rtc}, we also define the discretized inference delay
      as $d := \lfloor \tinfer/\Delta t_{\text{ctrl}}\rfloor$.

    \item \textit{Execution duration} $\texec := s \cdot \tctrl$. This denotes the time
      required for the robot client to execute $s$ actions.
  \end{itemize}

  \textbf{Synchronous Inference.} The system operates synchronously by default, as
  shown in \Cref{fig:infer_study}(a). After completing execution of the preceding
  chunk at time $t$, the client sends the observation $\mathbf{o}_{t}$ to the
  server to request a new inference. After the inference latency, the server
  returns the predicted chunk. During this period, the robot controller pauses
  and resumes only when the new actions arrive at $t + \tinfer$. To achieve uninterrupted
  execution, the condition $\tinfer < \Delta t_{\text{ctrl}}$ (\ie, $d=0$) should
  hold, meaning the next chunk is available within a single control step. In practice,
  however, this requirement is hardly satisfied, resulting in non-smooth
  trajectories and degraded task performance~\cite{rtc,vlash}.

  \textbf{Asynchronous Inference.} A natural strategy to tackle inter-chunk pauses
  is asynchronous inference~\cite{smolvla}. The core idea is to initiate
  inference of the next chunk before the current chunk is fully executed, as depicted
  in \Cref{fig:infer_study}(b). Specifically, once inference is triggered at
  time $t$, the robot continues executing the remaining actions in the ongoing chunk.
  By time $t + \tinfer$, when the final action is completed, the newly predicted
  chunk is expected to be available, thereby enabling seamless execution without
  halt.

  However, asynchronous execution incurs the problem of perception-execution gap~\cite{liao2025delay,
  dynamicvla}. The observation is captured at time $t$, but when the new chunk
  becomes available, the environment and the robot state may have changed due to
  actions executed during the interval $(t, t + \tinfer)$. A naive strategy of
  discarding the first $d$ delayed actions in the new chunk and switching to the
  remaining ones $[d, d+s)$ can lead to unstable and discontinuous motion, and
  this issue becomes increasingly severe as the delay $d$ grows~\cite{rtc, vlash}.
  Recent approaches mitigate inter-chunk discontinuity by incorporating the
  overlapping actions ($[s, d+s)$ in previous chunk) as part of the model
  input~\cite{rtc, trainingrtc, remac, legato, cai2026xiaomi}. This paper follows
  RTC~\cite{rtc, trainingrtc}, where the overlapping actions are treated as
  prefix conditions during action generation, guiding the new actions to transition
  smoothly.

  \begin{table}[t]
    \centering
    \caption{Reaction characteristics of synchronous and asynchronous inference.
    $\mathcal{U}(a, b)$ denotes a uniform distribution with lower and upper bounds
    $a$ and $b$, so the expectation of $\treact$ equals the midpoint $(a+b)/2$.}
    \footnotesize
    \setlength{\tabcolsep}{7pt}
    \begin{tabular}{c|cccc}
      \toprule Mode & Infer Interval   & $\treact\sim\dreact$                     & $\mathbb{E}[\treact]$    & $s_{\text{min}}$                              \\
      \midrule Sync & $\tinfer+\texec$ & $\mathcal{U}(\tinfer, 2*\tinfer+\texec)$ & $1.5*\tinfer+0.5*\texec$ & --                                             \\
      Async         & $\texec$         & $\mathcal{U}(\tinfer, \tinfer+\texec)$   & $\tinfer+0.5*\texec$     & $\lceil \tinfer/\Delta t_{\text{ctrl}}\rceil$ \\
      \bottomrule
    \end{tabular}
    \label{tab:react_study}
    \vspace{-5mm}
  \end{table}

  \textbf{Smoothness \vs Reaction.} Existing real-time VLAs primarily focus on improving
  inter-chunk smoothness. Nevertheless, they often overlook or misunderstand another
  fundamental aspect of real-time performance: \textbf{reaction}. In this work, we
  revisit the notion of reaction in action chunking policy inference and provide
  a systematic analysis.

  Reaction time $\treact$ is defined as the interval between the occurrence of a
  sudden event and the response produced by the robot.
  We summarize the reaction characteristics of synchronous and asynchronous
  clients in \Cref{tab:react_study}. A key insight is that reaction time is susceptible
  to the dual influence of inference latency and frequency. Given that policy
  inference is performed periodically, with consecutive inference triggers at time
  $t$ and $t'$ in \Cref{fig:infer_study}. When a new event occurs, the system can
  only respond after the next inference cycle is completed. Therefore, the lower
  bound of $\treact$ is $\tinfer$, corresponding to the case where the event
  happens just before an inference starts. In the worst case, the event occurs
  immediately after inference begins; the reaction will then only be reflected at
  $t' + \tinfer$, shaping an upper bound equal to $\tinfer$ plus the inference
  interval.

  As events occur stochastically in the physical world, the reaction time can be
  modeled as \textit{following a uniform distribution}, denoted by $\dreact$. An
  important finding from the expectation $\mathbb{E}[\treact]$ in
  \Cref{tab:react_study} is that the gain by upgrading from synchronous to
  asynchronous inference is potentially limited, as it reduces the expected reaction
  time by only $0.5*\tinfer$.

  Reducing the execution horizon $s$ is an intuitive idea to increase the inference
  frequency.
  In the asynchronous setting, $s$ has a minimum value $s_{\text{min}}:=\lceil \tinfer
  /\Delta t_{\text{ctrl}}\rceil$ to guarantee that the inference interval exceeds
  the latency (\ie, $\texec\ge\tinfer$). Under this configuration, inference is
  triggered every $s_{\text{min}}$ control steps, achieving optimal reaction performance~\cite{dynamicvla}.

  \noindent
  \textbf{Time to First Action.} As observed in \Cref{tab:react_study}, reaction
  time heavily depends on inference latency. More importantly, if actions are not
  generated simultaneously, responsiveness is determined solely by how quickly
  the system can produce the \textit{first} action. Since the robot does not
  need the entire action chunk to begin moving, later actions, while essential
  for task accuracy, do not directly affect immediate responsiveness. Therefore,
  we introduce \emph{Time to First Action (TTFA)} as a more precise metric for
  measuring reactivity, analogous to Time to First Token (TTFT) in large language
  models~\cite{holmes2024deepspeed, zhou2024survey}. TTFA explicitly captures
  the earliest moment at which the robot can initiate movement, making it the true
  bottleneck of reaction speed. This paper presents a novel asynchronous
  pipeline that jointly minimizes TTFA and increases inference frequency, leading
  to substantially improved reaction capability in action chunking policies.

  \section{Methodology}
  \label{sec:method}

  \subsection{Preliminaries}

  We adopt the widely used flow-based VLA structure~\cite{pi0,pi05, gr00t, xvla}.
  The model consists of a VLM backbone and an action expert (AE), learning a velocity
  field that transports a noise sample to the target action chunk using conditional
  flow matching~\cite{flowmatching, lipman2024flow}.
  Training follows the optimal transport formulation~\cite{tong2024improving, liu2023flow},
  which assumes a linear interpolation path between Gaussian noise $\bm{\epsilon}
  \sim \mathcal{N}(\mathbf{0}, \mathbf{I})$ and the ground-truth actions $\hat{\A}
  _{t}$: $\A_{t}^{\tau}= \tau \bm{\epsilon}+ (1 - \tau)\hat{\A}_{t},$ where
  $\tau\in (0,1)$ is the continuous timestep of the flow. The objective is to regress
  the velocity field along this path with network $v_{\theta}$:
  \begin{equation}
    \mathcal{L}(\theta)=\mathbb{E}_{\tau\sim\mathcal{U}(0,1)}\left\|v_{\theta}(\mathbf{o}
    _{t}, \A_{t}^{\tau}, \tau)-(\bm{\epsilon}-\hat{\A}_{t}) \right\|^{2}. \label{eq2}
  \end{equation}

  During inference, actions are generated by initializing from Gaussian noise
  $\A_{t}^{1}\sim \mathcal{N}(\mathbf{0}, \mathbf{I})$ at $\tau = 1$, and progressively
  integrating the learned velocity field toward $\tau = 0$ using an ODE solver such
  as the Euler method:
  \begin{equation}
    \A_{t}^{\tau+\Delta\tau}=\A_{t}^{\tau}+v_{\theta}(\mathbf{o}_{t}, \A_{t}^{\tau}
    , \tau)\Delta \tau, \label{eq3}
  \end{equation}
  where $\Delta\tau = -1/N$ is related to the number of sampling steps $N$, a typical
  value of 10 in practice.

  \subsection{Pilot Study on Action Chunk Sampling}
  \label{sec:pilot_study}

  \begin{figure}[t]
    \centering
    \begin{subfigure}
      {0.3\linewidth}
      \centering
      \includegraphics[width=0.95\linewidth]{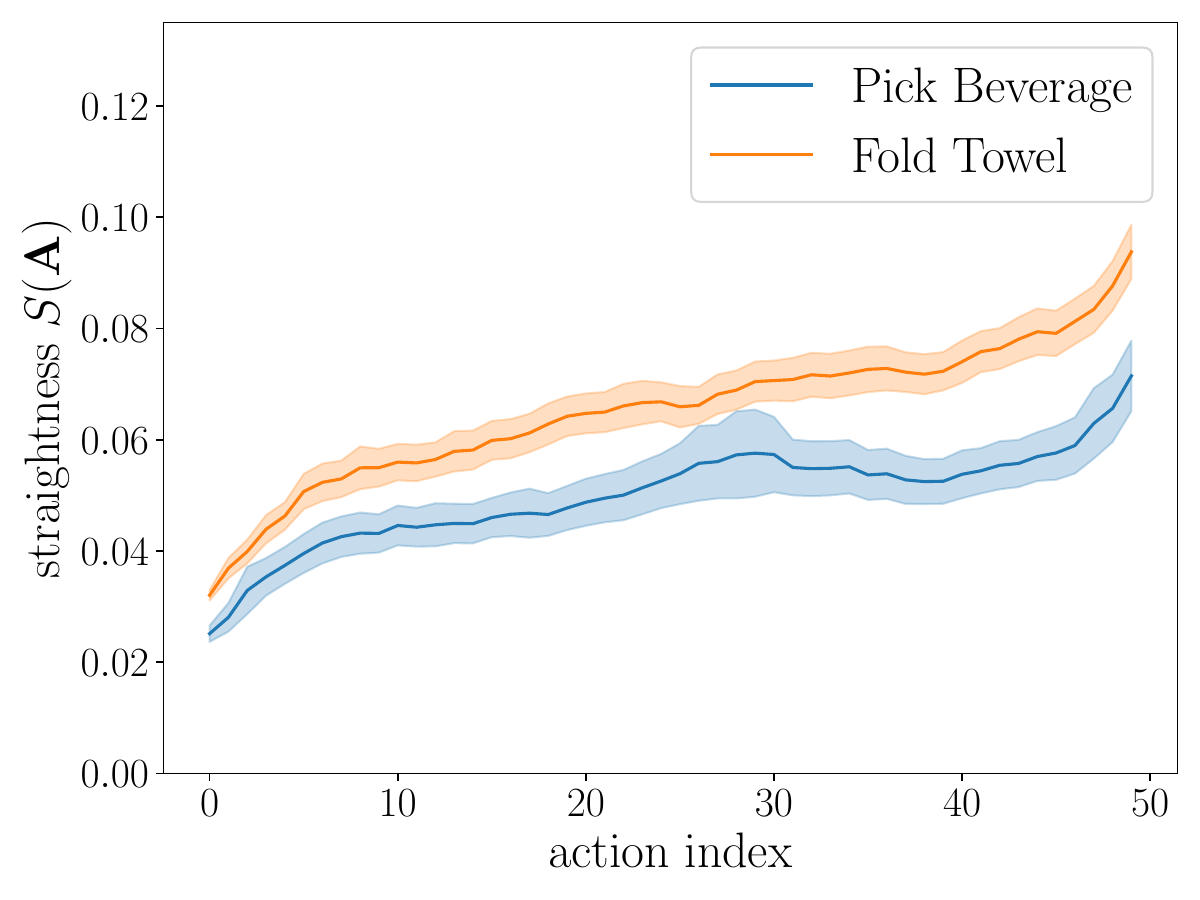}
      \vspace{-2mm}
      \caption{}
      \label{fig:pilot_study_stra}
    \end{subfigure}
    \begin{subfigure}
      {0.3\linewidth}
      \centering
      \includegraphics[width=0.95\linewidth]{
        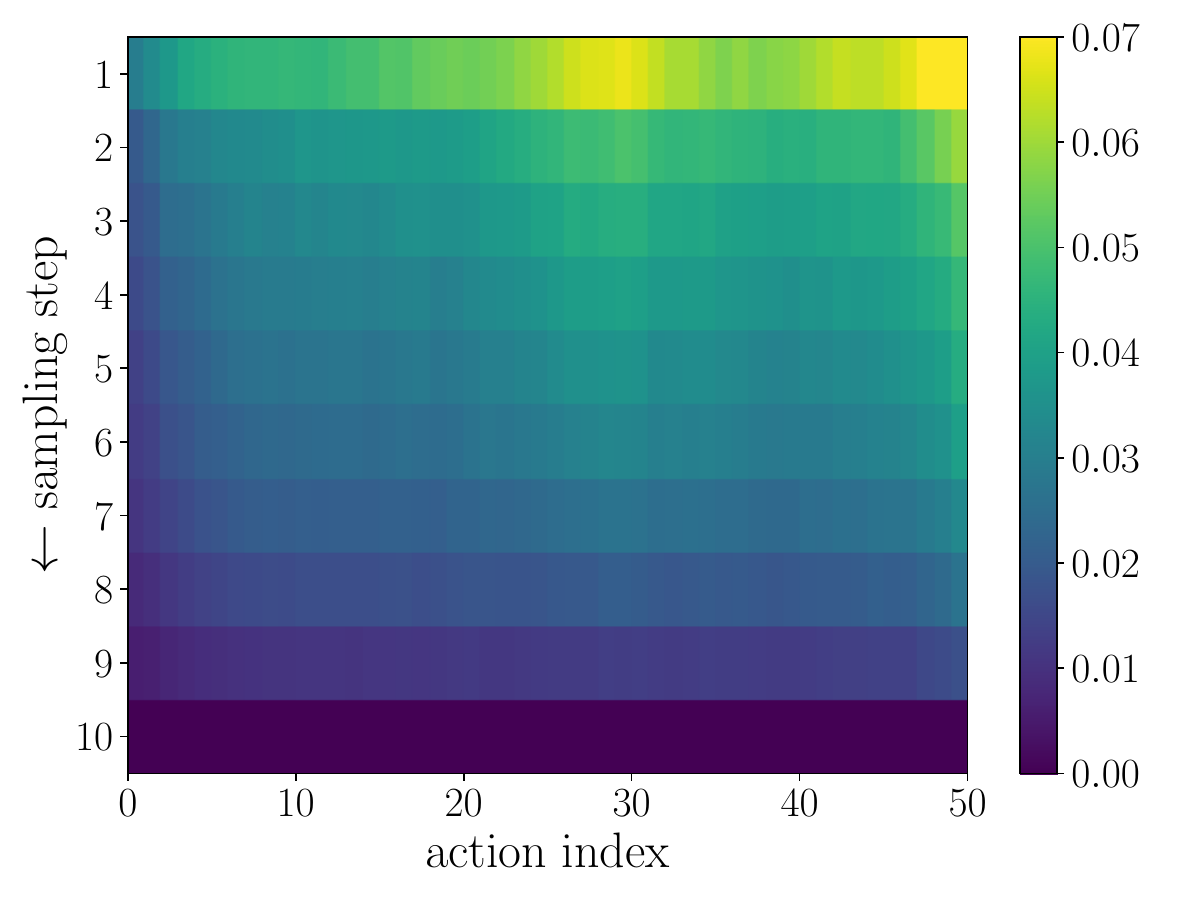
      }
      \vspace{-2mm}
      \caption{}
      \label{fig:pilot_study_x0_bev}
    \end{subfigure}
    \begin{subfigure}
      {0.3\linewidth}
      \centering
      \includegraphics[width=0.95\linewidth]{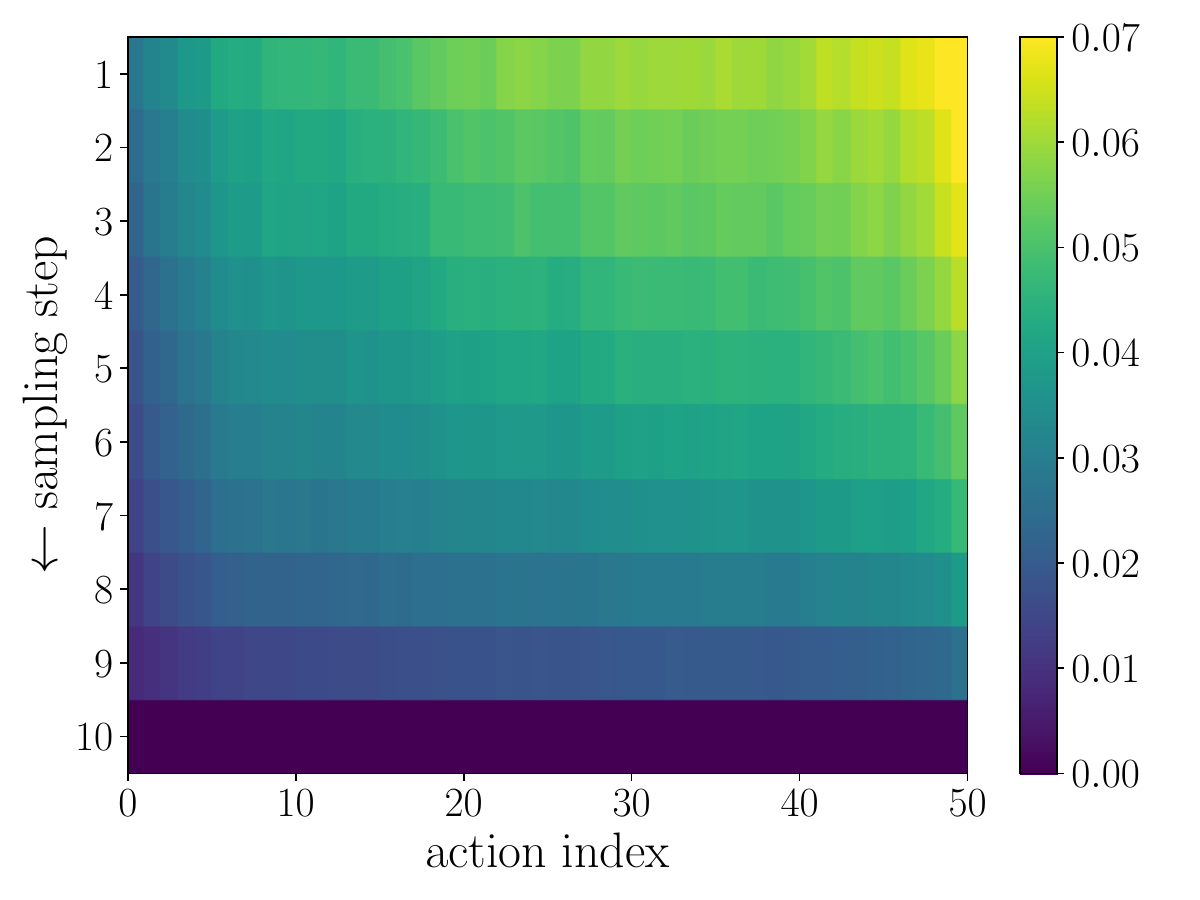}
      \vspace{-2mm}
      \caption{}
      \label{fig:pilot_study_x0_towel}
    \end{subfigure}
    \vspace{-1mm}
    \caption{Visualizations of (a) straightness $S(\A)$ of the denoising path
    during sampling of the action chunk, and (b,c) differences between the intermediate
    clean action estimates $\tilde{\A}_{t}^{\tau\rightarrow 0}$ at each sampling
    timestep $\tau$ and the final output $\A_{t}^{0}$, using the tasks ``Pick Beverage''
    and ``Fold Towel'', respectively.}
    \label{fig:pilot_study}
    \vspace{-5mm}
  \end{figure}

  Existing flow-based VLAs treat the entire action chunk as an indivisible unit
  and apply a constant timestep schedule across all action indexes. As a result,
  every action within the chunk undergoes the same number of denoising steps
  during inference. The immediate next action $\A_{t}$, which is urgently
  required for execution, is therefore forced to share the same schedule as the most
  distant future action $\A_{t+H-1}$. Consequently, the entire multi-step denoising
  procedure has to be completed before any individual action can be issued,
  which constitutes a dominant bottleneck of the overall latency~\cite{yang2025efficientvla}.

  Nevertheless, action chunks exhibit an inherent temporal structure. Given
  current observations and proprioceptive states, early-stage actions are subject
  to stronger causal constraints and thus lie in a substantially narrower search
  space compared to future actions. Intuitively, this makes short-term predictions
  easier and more certain. Furthermore, when asynchronous methods incorporate action
  prefixes as input~\cite{trainingrtc, legato, vlash}, these extra priors
  provide additional conditioning that constrains subsequent predictions. This further
  reduces the uncertainty of the immediate actions and lowers the complexity of generation.

  We validate this hypothesis through a quantitative analysis of the sampling dynamics
  in flow-based VLAs. Specifically, we adopt the \emph{straightness} metric~\cite{liu2023flow},
  which is defined for any continuously differentiable process $\Z=\{Z_{\tau}\}$
  evolving from $Z_{0}$ to $Z_{1}$ as $S(\Z) = \int_{0}^{1}\mathbb{E}\left[ \| (Z
  _{1}- Z_{0}) - \dot{Z}_{\tau}\|^{2}\right ] \mathrm{d}\tau,$ where $\dot{Z}_{\tau}
  =\frac{\mathrm{d}}{\mathrm{d}\tau}Z_{\tau}$ denotes the instantaneous velocity
  at time $\tau$. In our context, the VLA denoising process is discretized and the
  straightness can be formulated as:
  \begin{equation}
    S(\A)=\sum_{\tau=0}^{1}\mathbb{E}_{t}\left[ \left\| (\A_{t}^{1}- \A_{t}^{0})
    - v_{\theta}(\mathbf{o}_{t}, \A_{t}^{\tau}, \tau)\right\|^{2}\right] \cdot (-
    \Delta \tau) ,
  \end{equation}
  where $\A_{t}^{0}$ represents the final actions obtained via \Cref{eq3}. A value
  of $S(\A)=0$ indicates a perfectly straight path. Smaller $S(\A )$ corresponds
  to paths closer to linear interpolation, which in turn can be accurately integrated
  with fewer steps~\cite{liu2023flow}. We also investigate the estimated clean
  actions $\tilde{\A}_{t}^{\tau\rightarrow 0}$ at each denoising step
  $\{1, 1+\Delta \tau, \dots, -\Delta \tau\}$, obtained via the following extrapolation:
  \begin{equation}
    \tilde{\A}_{t}^{\tau\rightarrow 0}=\A_{t}^{\tau}-v_{\theta}(\mathbf{o}_{t}, \A
    _{t}^{\tau}, \tau)\tau.
  \end{equation}
  We measure their deviation from the final output $\A_{t}^{0}$ using $\ell_{2}$
  norm $\|\tilde{\A}_{t}^{\tau\rightarrow 0}- \A_{t}^{0}\|_2$. During sampling,
  this deviation is expected to decrease and reaches zero at the
  final step. A smaller deviation suggests that the model provides a more accurate
  estimate of the results at current timestep.

  We conduct a pilot study by fine-tuning a pretrained $\pi_{0.5}$ model on our real-world
  robotic tasks. As visualized in \Cref{fig:pilot_study}, we can find that both the
  straightness metric and the estimate deviation exhibit non-uniformity across the
  temporal dimension (action index) of the action chunk. In particular, early actions
  (approximately the first 1--10 frames) demonstrate lower straightness values
  and smaller variations in $\tilde{\A}_{t}^{\tau\rightarrow0}$ throughout the sampling
  iterations. This empirical observation provides strong evidence supporting our
  hypothesis.

  \subsection{FASTER}

  Motivated by the insight that near-term actions within a chunk are easier to generate
  under flow matching, we propose to prioritize the sampling of these latency-critical
  actions with a Horizon-Aware Schedule (HAS).

  \begin{figure}[t]
    \centering
    \includegraphics[width=0.8\linewidth]{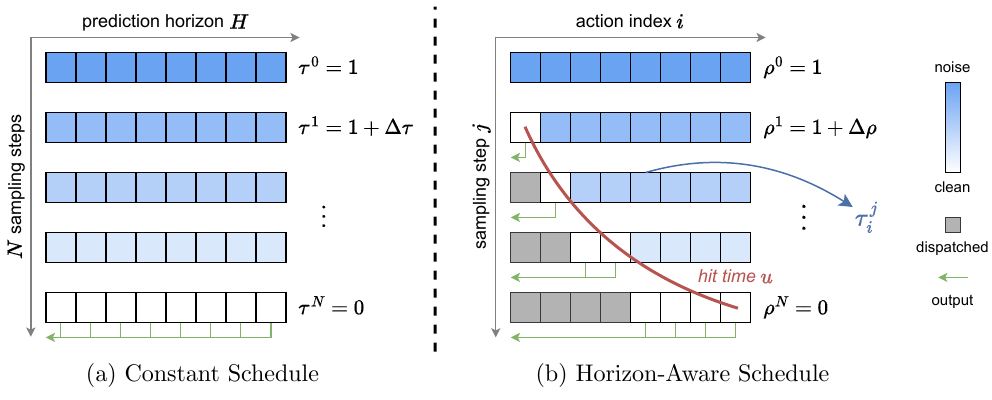}
    \vspace{-3mm}
    \caption{Illustration of (a) constant timestep schedule used in conventional
    flow sampling and (b) Horizon-Aware Schedule (HAS) used in FASTER that
    allocates adaptive hit times across the action chunk and accelerates the
    sampling of early actions, enabling streaming output.}
    \label{fig:schedule}
    \vspace{-5mm}
  \end{figure}

  \noindent
  \textbf{Horizon-Aware Schedule.} Unlike conventional flow-based VLAs, which employ
  a constant time schedule across the entire chunk (\Cref{fig:schedule}(a)),
  we design a horizon-aware time allocation mechanism that accelerates the denoising
  of near-term actions during inference, while allowing later-horizon actions to
  follow a comparatively slower schedule, mimicking the straightness plot in \Cref{fig:pilot_study}(a).

  Inspired by Diffusion Forcing~\cite{diffusionforcing}, the flow matching timestep
  is made index-dependent and represented as a vector $\bm{\tau}=\{{\tau}_{i}\}$,
  where $i\in [0, H-1]$ denotes the action index. As illustrated in \Cref{fig:schedule}(b),
  we use $\rho^{j}\in (0, 1)$ to represent the global sampling progress at $j$-th
  step, which still iterates from 1 to 0 following the standard flow matching
  procedure. Each action reaches completion at a distinct \emph{hit time}
  ${u}_{i}$ determined by its index:
  \begin{equation}
    {u}_{i}= \left(1-(i/(H-1))^{\alpha}\right)\cdot{u}_{0}\quad i\in [1, H-1], \label{eq5}
  \end{equation}
  where the predefined ${u}_{0}$ specifies the global timestep at which the first
  action is finalized. The hyperparameter $\alpha\in (0, 1]$ controls how the
  hit times vary across the action index. When $\alpha=1$, the hit times
  decrease uniformly from $u_{0}$ to 0. When $\alpha<1$, early actions reach their
  hit times rapidly, while future actions have hit times closer to 0. This
  design adaptively allocates more denoising steps to future actions due to
  compounding uncertainty, preserving generation quality and minimizing deviation
  from the original schedule inherited from pretraining.

  Given the global timestep $\rho^{j}$, the local time schedule for each action
  index is formulated as:
  \begin{equation}
    \tau_{i}^{j}= \max \left(0, (\rho^{j}-u_{i})/(1-u_{i})\right).
  \end{equation}
  Under this schedule, once $\rho^{j}$ reaches ${u}_{0}$, the first action is fully
  denoised and can be dispatched immediately, while the remaining actions
  continue refining. We set ${u}_{0}=(N-1)/N$, thus guaranteeing that the first
  action is ready with a single sampling step. With $\rho^{j}$ further decreasing,
  subsequent actions are progressively completed when $\rho^{j}={u}_{i}$.

  \noindent
  \textbf{Fine-tuning with Mixed Schedule.} Directly fine-tuning a pretrained VLA
  using HAS may introduce two challenges. First, existing pretrained models are optimized
  under a constant timestep schedule, thus naively switching to an index-dependent
  schedule can enlarge the fine-tuning gap, in addition to the distribution shift.
  Second, when randomly selecting $\rho\sim\mathcal{U}(0, 1)$, there exists a
  high probability that the corresponding local timestep for near-term actions
  becomes zero. This could collapse the learning of these actions since the inputs
  are constantly ground-truth and potentially deteriorating rollout performance.

  To address these issues, we introduce a mixed scheduling strategy to augment fine-tuning.
  Concretely, given a mixing probability $p$, each action sample in a training batch
  utilizes HAS with probability $p$, and retains the original constant schedule
  with probability $1-p$. The mixed schedule forces the model to learn the flow
  matching velocity field under both timestep parameterizations, thereby improving
  robustness to the schedule variation across the action horizon. It is worth noting
  that the proposed scheduling methodology can be readily incorporated into the standard
  fine-tuning pipeline of flow-based VLAs, without any architecture
  modifications or additional training cost.

  \noindent
  \textbf{Synergization with Action Conditioning.} HAS naturally synergizes with
  the action conditioning technique proposed in Training-time RTC~\cite{trainingrtc},
  which conditions both training and sampling on the action prefixes by treating
  them as fully denoised. Under our adaptive schedule, early actions frequently
  receive timesteps close to zero, while future actions are assigned progressively
  larger timesteps. This pattern is inherently consistent with the principle of
  action conditioning. Moreover, it encourages the model to learn a more
  structured mapping between timestep values and the degree of noise
  interpolation relative to the target actions, thereby strengthening its understanding
  of temporally conditioned generation. Action conditioning is integrated into
  HAS with an offset on the index, and we provide a detailed description in Appendix~\ref{supp:has_ac}.

  \noindent
  \textbf{Streaming Client-Server Interface.} With \method, actions are generated
  progressively during the denoising iterations. To exploit this property in
  real-world robotic systems, we implement a streaming client-server interface.
  On the server side, newly finalized actions from the policy are dispatched immediately,
  while the model proceeds to generate the remaining actions concurrently. On the
  client side, the controller continuously listens for incoming packets and
  appends received actions to the robot’s action buffer for execution, without waiting
  for the entire chunk to complete. As long as the action acquisition rate
  exceeds the robot control frequency---which is achievable even on consumer-level
  GPUs---the robot can operate without interruption (shown in Appendix~\ref{supp:stream}).

  \input{table2}

  \noindent
  \textbf{Enhancing Reaction Capability.} We analyze how \method improves reaction
  time by tightening both its lower and upper bounds. For conventional flow-based
  VLAs with $N$ sampling steps, TTFA is approximately $\Delta t_{\text{VLM}}+N\Delta
  t_{\text{AE}}$, where $\Delta t_{\text{VLM}}$ and $\Delta t_{\text{AE}}$
  denote the inference time of the VLM and AE, respectively. Equipped with
  \method, TTFA is shortened to $\Delta t_{\text{VLM}}+\Delta t_{\text{AE}}$, as
  the first action requires only a single AE sampling step.

  Responsiveness can be further improved by increasing the inference frequency, \ie,
  reducing the execution horizon $s$, whose minimum value $s_{\text{min}}$ must satisfy
  $\texec\ge\tinfer$. When a high frequency is desired, a relatively small $s$ is
  selected (\eg 4 out of $H=50$), implying that only a small portion of the
  chunk is useful while the other actions are discarded. Our method enables an early-stopping
  strategy: once all actions within the execution horizon are finalized, the remaining
  sampling steps can be skipped. This avoids completing all AE iterations and
  effectively reduces the overall latency. Consequently, we can apply a smaller
  feasible $s_{\text{min}}$ compared with conventional asynchronous inference, tightening
  the upper bound of $\treact$ without sacrificing execution smoothness.

  \section{Experiments}
  \label{sec:exp}

  We focus primarily on real-world experiments, simulation benchmarks and ablation studies are provided in Appendix~\ref{supp:sim} and \ref{supp:abla} due to space limitations.

  \subsection{Experimental Analysis on Reaction Speed}

  \textbf{Experimental Setup.} To investigate improvements in reaction capability,
  we first compare the inference latency---measured by TTFA---and the expected
  reaction time of our method against the synchronous and asynchronous baselines.
  Since prior real-time VLA approaches~\cite{rtc,trainingrtc,vlash} follow the same
  naive asynchronous paradigm in these aspects, we do not report them separately.
  Experiments are conducted on two hardware platforms: a high-performance RTX
  4090 GPU, and a consumer-grade RTX 4060 GPU. We utilize two representative
  flow-based VLAs, $\pi_{0.5}$~\cite{pi05} and X-VLA~\cite{xvla}, and adhere to
  their default configurations. The robot control frequency is set to $f=30$Hz, corresponding
  to a control period $\tctrl=33.3$ms. To reflect the optimal achievable reaction
  speed, we set the execution horizon to $s_{\text{min}}$, thereby maximizing the
  inference frequency.

  \noindent
  \textbf{Results.} As detailed in \Cref{tab:react}, our method manifests substantial
  acceleration in reaction performance across all scenarios. As the design of X-VLA
  incurs a higher computational cost in its action expert, \method delivers particularly
  significant gains, achieving a 3$\times$ boost in TTFA on RTX 4060. Notably,
  the early-stopping strategy in our approach effectively decreases the feasible
  inference interval by a smaller $s_{\text{min}}$, which minimizes the inference-execution
  cycle and contributes to additional enhancements in responsiveness.

  As highlighted in \Cref{sec:analysis}, reaction time should be treated as a
  random variable, since event occurrences are inherently stochastic.
  Accordingly, \Cref{tab:react_prob} presents a probabilistic analysis that more
  faithfully reflects real-world conditions, measuring the probability that one
  method attains a faster reaction time than another. \method surpasses both baselines
  by a clear margin, with a larger advantage in the more resource-constrained
  scenario. Notably, on X-VLA, our method is deterministically superior: its
  upper bound of reaction time is lower than the baselines' lower bound,
  establishing a strict performance dominance.

  \input{table3}

  \subsection{Real-world Experiments}

  \textbf{Experimental Setup.} To examine the real-robot reaction capability in a
  highly dynamic environment, we design a task that demands both rapid response
  and accurate motion execution. Specifically, we train VLA to play table tennis
  using a racket mounted on a 6-DoF Piper robot arm in the AgileX Cobot Magic
  platform. We collect approximately 14 minutes of demonstration data via human teleoperation
  to fine-tune the $\pi_{0.5}$ models. In addition, we include two tasks that place
  less emphasis on real-time reaction: (1) ``Pick Beverage'': picking up a beverage
  and placing it in a basket, focusing on object and localization generalization;
  (2) ``Fold Towel'': folding a towel twice with dual arms, representing deformable
  object manipulation. Details are provided in Appendix~\ref{supp:real}.

  \begin{figure}[t]
    \centering
    \includegraphics[width=\linewidth]{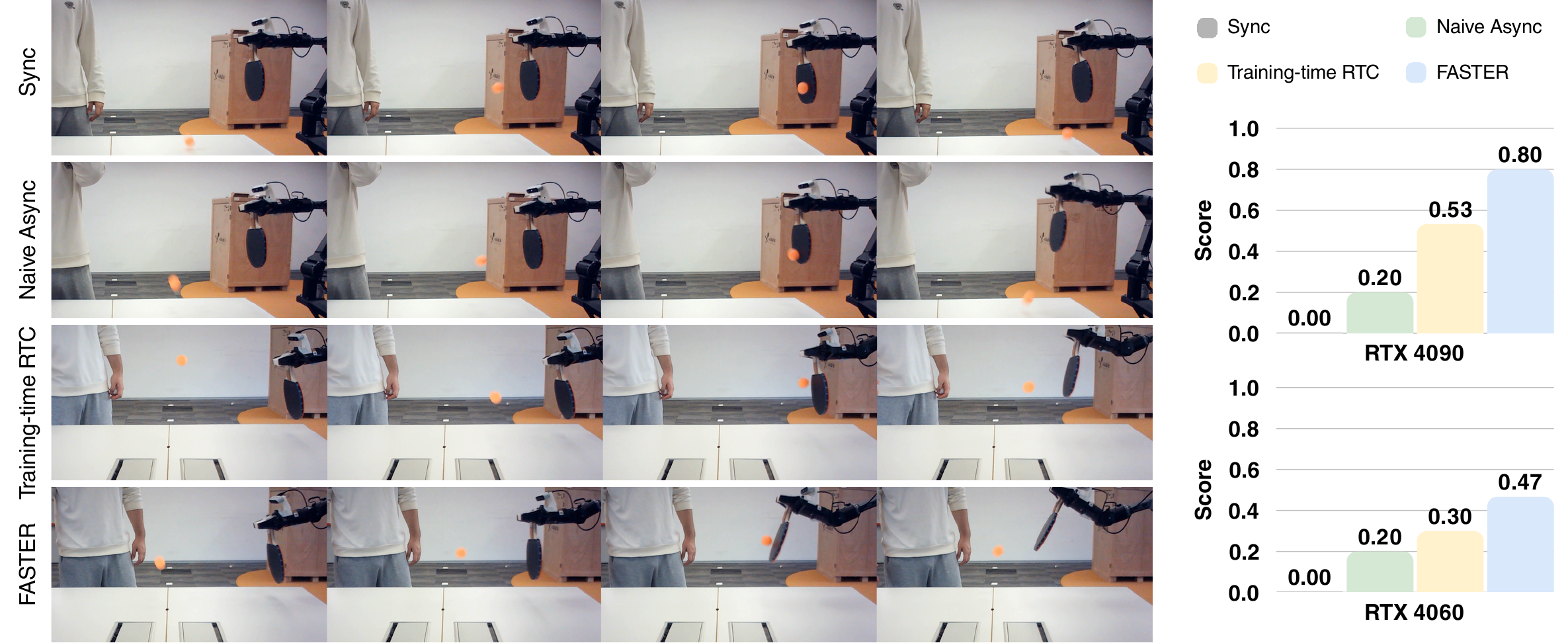}
    \vspace{-5mm}
    \caption{Comparison of real-world reaction speed on the table tennis task. \textbf{Left}:
    Visualization of rollouts on RTX 4090, the third column corresponds to the
    contact moment, and the interval between each image in a row is 166.7ms (5 frames). \textbf{Right}:
    Quantitative completion scores on two GPUs. }
    \label{fig:pp}
    \vspace{-3mm}
  \end{figure}

  \begin{figure}[t]
    \centering
    \includegraphics[width=0.9\linewidth]{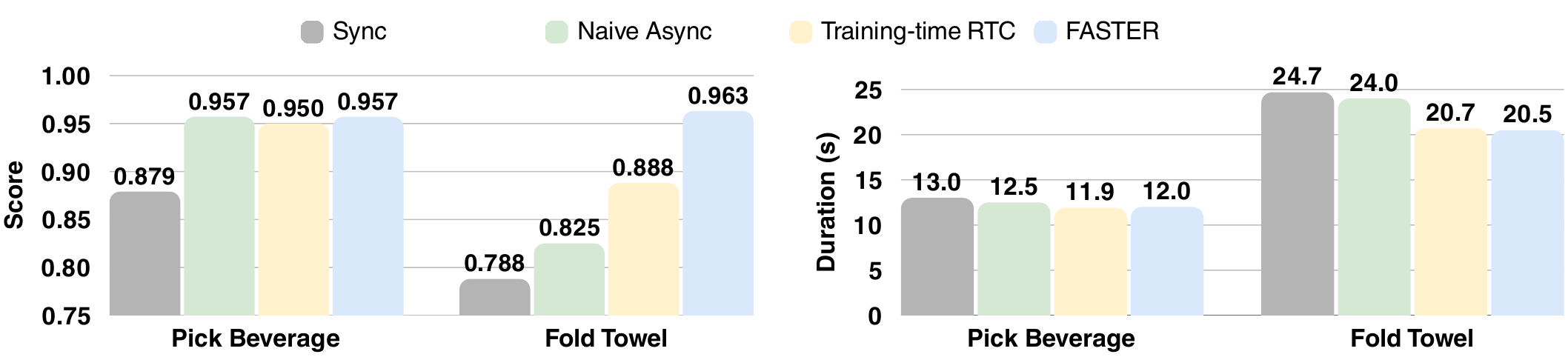}
    \vspace{-2mm}
    \caption{Comparison of real-world performance and completion duration on two
    additional tasks.}
    \label{fig:exp_tasks}
    \vspace{-5mm}
  \end{figure}

  \noindent
  \textbf{Results.} We compare synchronous inference (``Sync''), naive
  asynchronous inference (``Naive Async''), and the state-of-the-art Training-time
  RTC~\cite{trainingrtc} with our \method. Representative rollouts are
  visualized on the left of \Cref{fig:pp}, with quantitative results presented in
  the right panels. We observe that Sync fails to respond to
  incoming balls in our trials due to its drastically slow reaction speed. Naive Async and Training-time
  RTC share the same temporal pipeline and thus similar reaction capability.
  However, Training-time RTC improves the smoothness of racket swing through its
  action conditioning mechanism, leading to moderately higher scores.

  Our method exhibits noticeably faster reaction than all baselines, which can be
  intuitively observed from the racket angle at the moment of ball contact. To
  successfully return the ball, the robot should begin adjusting its arm posture
  in advance to reach an appropriate hitting position with sufficient swing velocity.
  If the reaction is delayed, there is insufficient time left to move, leading
  to suboptimal contact angles, as seen in the baselines. In contrast, faster reactions
  allow the robot to initiate motion earlier, providing adequate time to rotate
  the racket and build up swing speed, resulting in a more powerful and controlled
  hit. Therefore, the racket angle serves as a clear and visually interpretable indicator
  of reaction speed. Overall, \method achieves the best performance across both hardware
  settings. The advantage is particularly pronounced on RTX 4060, where inference
  runs at merely 3Hz, highlighting the compounded benefits of reduced inference
  latency and increased frequency.

  We report the average completion score and duration for the additional tasks
  in \Cref{fig:exp_tasks}. 
  The results show that asynchronous methods outperform
  Sync by a clear gap, and \method achieves better or comparable scores across
  both tasks. 
  This highlights that task performance is not determined
  solely by action accuracy, but also by the real-time interaction with the
  physical world. Though our method may slightly compromise action prediction due
  to accelerated sampling, it strikes a more effective balance between
  responsiveness and accuracy.

  Consistent with previous findings~\cite{trainingrtc, vlash}, the synchronous method
  exhibits the longest task duration due to frequent inter-chunk pauses. In
  contrast, Training-time RTC and \method effectively reduce completion time,
  yielding greater efficiency for downstream applications.

  \section{Conclusion}
  \label{sec:con}

  In this paper, we revisit reaction capability in action chunking VLA policies
  and identify the constant timestep schedule in flow-based VLAs as a key
  bottleneck of real-time responsiveness. We propose FASTER, which leverages a
  Horizon-Aware Schedule to adaptively accelerate action sampling. It enables single-step
  generation of the immediate action, without compromising overall trajectory
  quality. Capitalizing on a streaming client-server interface with early stopping,
  FASTER jointly reduces TTFA and speeds up closed-loop control. Real-robot experiments
  confirm that FASTER offers a robust, general, and plug-and-play path toward real-time
  embodied intelligence, particularly on edge devices.

  {\small \bibliographystyle{plain} \bibliography{main} }

  \newpage
  \appendix
  \input{supp}

\end{document}

%% file: table2.tex
  \begin{table}[t]
    \centering
    \caption{Comparison of reaction capability on RTX 4090 and RTX 4060 GPUs. ``\textit{Speedup}''
    denotes the relative gain of our method over the asynchronous baseline.}
    \footnotesize
    \setlength{\tabcolsep}{8pt}
    \begin{tabular}{cc|cccccccc}
      \toprule \multirow{2}{*}{Model}       & \multirow{2}{*}{Method} & \multicolumn{3}{c}{RTX 4090}                & \multicolumn{3}{c}{RTX 4060}              \\
                                            &                         & TTFA $\downarrow$                           & $s_{\min}$ $\downarrow$                  & $\mathbb{E}[\treact]$ $\downarrow$          & TTFA $\downarrow$                           & $s_{\min}$ $\downarrow$                     & $\mathbb{E}[\treact]$ $\downarrow$          \\
      \midrule \multirow{4}{*}{$\pi_{0.5}$} & Sync                    & 80.0$_{\pm 1.6}$ms                          & 3                                        & 170.0ms                                     & 303.3$_{\pm 0.8}$ms                         & 10                                          & 621.6ms                                     \\
                                            & Async                   & 80.0$_{\pm 1.6}$ms                          & 3                                        & 130.0ms                                     & 303.3$_{\pm 0.8}$ms                         & 10                                          & 470.0ms                                     \\
                                            & \method                 & \textbf{62.1}$_{\pm 3.1}$ms                 & 3                                        & \textbf{112.1}ms                            & \textbf{238.6}$_{\pm 1.9}$ms                & \textbf{8}                                  & \textbf{371.9}ms                            \\
                                            & \textit{Speedup}        & \cellcolor{mygreen!80}\textbf{1.29$\times$} & \cellcolor{mygreen!80}\textbf{--}        & \cellcolor{mygreen!80}\textbf{1.16$\times$} & \cellcolor{mygreen!80}\textbf{1.27$\times$} & \cellcolor{mygreen!80}\textbf{1.25$\times$} & \cellcolor{mygreen!80}\textbf{1.26$\times$} \\
      \midrule \multirow{4}{*}{X-VLA}       & Sync                    & 113.7$_{\pm 0.8}$ms                         & 4                                        & 237.2ms                                     & 399.5$_{\pm 8.5}$ms                       & 12                                          & 799.2ms                                     \\
                                            & Async                   & 113.7$_{\pm 0.8}$ms                         & 4                                        & 180.4ms                                     & 399.5$_{\pm 8.5}$ms                       & 12                                          & 599.5ms                                     \\
                                            & \method                 & \textbf{44.8}$_{\pm 0.3}$ms                 & \textbf{2}                               & \textbf{78.1}ms                             & \textbf{129.2}$_{\pm 2.4}$ms              & \textbf{6}                                  & \textbf{229.2}ms                            \\
                                            & \textit{Speedup}        & \cellcolor{mygreen!80}\textbf{2.54$\times$} & \cellcolor{mygreen!80}\textbf{2$\times$} & \cellcolor{mygreen!80}\textbf{2.31$\times$} & \cellcolor{mygreen!80}\textbf{3.09$\times$} & \cellcolor{mygreen!80}\textbf{2$\times$}    & \cellcolor{mygreen!80}\textbf{2.62$\times$} \\
      \bottomrule
    \end{tabular}
    \label{tab:react}
    \vspace{-7mm}
  \end{table}

%% file: table3.tex
  \begin{table}[t]
    \centering
    \caption{Comparison of reaction speed from a probabilistic perspective on
    RTX 4090 and RTX 4060 GPUs. ``\vs Sync'' denotes the probability that a given
    method has a shorter reaction time than Sync, and ``\vs Async'' is defined
    analogously.}
    \footnotesize
    \setlength{\tabcolsep}{10pt}
    \begin{tabular}{cc|cccccc}
      \toprule \multirow{2}{*}{Model}       & \multirow{2}{*}{Method} & \multicolumn{2}{c}{RTX 4090} & \multicolumn{2}{c}{RTX 4060} \\
                                            &                         & \vs Sync                     & \vs Async                   & \vs Sync      & \vs Async     \\
      \midrule \multirow{2}{*}{$\pi_{0.5}$} & Async                   & 0.72                         & -                           & 0.74          & -             \\
                                            & \method                 & \textbf{0.81}                & \textbf{0.66}               & \textbf{0.88} & \textbf{0.77} \\
      \midrule \multirow{2}{*}{X-VLA}       & Async                   & 0.73                         & -                           & 0.75          & -             \\
                                            & \method                 & \textbf{1.00}                & \textbf{1.00}               & \textbf{1.00} & \textbf{1.00} \\
      \bottomrule
    \end{tabular}
    \label{tab:react_prob}
    \vspace{-5mm}
  \end{table}

%% file: supp.tex
\begin{center}
  {\Large \textbf{Appendix}}
\end{center}

We provide the following contents in the appendix:
\begin{itemize}[noitemsep, topsep=0pt, leftmargin=*]
  \item \Cref{supp:related} reviews related work on VLA models, real-time VLAs, diffusion
    acceleration approaches in VLAs, and discusses closely related directions.

  \item \Cref{supp:async} provides a fine-grained analysis of the asynchronous inference
    pipeline and clarifies the relationship between inference latency, delay, and
    minimal execution horizon.

  \item \Cref{supp:study} presents additional pilot study results that further support
    the non-uniform sampling difficulty across action indices.

  \item \Cref{supp:method} gives additional methodological details, including
    the integration with action conditioning, the streaming client-server interface,
    and pseudo-code for training and inference.

  \item \Cref{supp:imple} summarizes implementation details for real-world experiments,
    simulation benchmarks, policy training, and robot deployment.

  \item \Cref{supp:exp} reports additional experimental results, including
    reaction analysis, extra real-world results, simulation benchmark results, and
    ablation studies.

  \item \Cref{supp:limitation} discusses the limitations of this work and potential
    directions for future research.

  \item \Cref{supp:impact} discusses potential positive and negative societal impacts.
\end{itemize}

\section{Related Work}
\label{supp:related}

\textbf{Vision-Language-Action Models.} Vision-Language-Action~(VLA) models~\cite{openvla,pi0,gr00t,drivepi,mimo}
extend large-scale vision-language pretraining from Vision-Language Models~(VLMs)~\cite{bai2023qwen,yang2025qwen3,li2024llava}
to embodied action learning, and have demonstrated impressive performance in
robotic manipulation. By pretraining on large-scale vision-text-action corpora~\cite{agibot_world,
oxe, bridgedata, droid}, VLAs enable robots to map multimodal observations and
language instructions directly to low-level motor commands, facilitating dexterous
manipulation across a wide range of tasks and promoting generalization to diverse
and complicated environments~\cite{shi2025diversity,pi05,pi06, fan2026any3d,
geminirobotics}.

Early approaches such as RT-2~\cite{rt2} and OpenVLA~\cite{openvla} discretize robot
actions into tokens, making them compatible with the auto-regressive objective
of VLMs. Subsequent effort explores diffusion- or flow-matching-based action
generation~\cite{dp}, adopting continuous action representations to model the multimodal
distribution. Represented by methods including $\pi_{0}$~\cite{pi0} and GR00T~\cite{gr00t},
these approaches incorporate a dedicated action expert alongside the VLM backbone,
generating high-quality actions conditioned on vision-language features.

\noindent
\textbf{Real-Time VLAs.} In contrast to VLMs operating purely in cyberspace,
VLAs interact with the physical world and are therefore highly sensitive to real-time
interaction~\cite{li2025train, dynamicvla}. Consequently, improving the
efficiency of VLAs has become an active research topic~\cite{yu2025survey, guan2025efficient}.
A straightforward strategy is to reduce model inference latency, with
representative directions including:
\begin{itemize}[noitemsep, topsep=0pt, leftmargin=*]
  \item Replacing the VLM backbone (typically $3\sim 7$B parameters) with
    smaller models consisting of fewer than 1B parameters~\cite{xvla,tinyvla,ni2025swiftvla,lin2025evo,chen2025nanovla,xiong2025hypervla,reuss2025flower,smolvla,liu2024robomamba,vlaadapter,budzianowski2025edgevla};

  \item Compressing the LLM backbone through mechanisms such as layer selection and
    early exiting~\cite{zhang2025mole,yang2025efficientvla,smolvla,chen2025rlrc,yue2024deer,jeon2026shallow,ye2025actdistill,jabbour2025don,song2025ceed,
    gr00t, yu2026ac};

  \item Accelerating action decoding, mainly for auto-regressive VLAs~\cite{wang2025spec,song2025ceed,oft,song2025accelerating,pertsch2025fast};

  \item Pruning visual tokens, as multi-view image inputs account for a large
    proportion of tokens while often introducing perceptual redundancy~\cite{ye2025token,liu2025vla,gao2025compressor,li2025semanticvla,pei2025action,fang2025sqap,wang2025specprune,jiang2025better,li2025sp,
    vlacache,huang2026environment, wei2026learning,yu2026ac};

  \item Applying low-level inference optimizations or quantization techniques~\cite{dai2025actionflow,ma2025blurr,williams2025lite,ma2025running,taherin2025cross,fang2025sqap,wu2025device,chen2025rlrc,wang2025bitvla,park2025saliency,park2024quantization}.
\end{itemize}

Another line of work seeks to eliminate inter-chunk pauses introduced by
standard action chunking and synchronous inference paradigm. By introducing asynchronous
execution, VLA models can generate the next action chunk concurrently while the
current one is being executed, resulting in non-stop trajectories~\cite{smolvla,
vla-rail, dynamicvla, cai2026xiaomi,legato}. However, naively switching between chunks
may cause abrupt multimodal transitions and jerky motions, a phenomenon known as
inter-chunk discontinuity~\cite{rtc,liao2025delay}. To mitigate this issue, RTC~\cite{rtc}
inpaints the next action chunk conditioned on the current chunk, while Training-time
RTC~\cite{trainingrtc}, REMAC~\cite{remac}, and VLASH~\cite{vlash} condition the
model on the predicted actions.

Different from prior work, \method is the first real-time VLA that explicitly
targets responsiveness by accelerating the sampling of immediate actions. Notably,
it requires no architectural modifications, rendering it orthogonal and complementary
to aforementioned efficient VLA techniques, such as LLM compression and token
pruning.

\noindent
\textbf{Diffusion Acceleration in VLAs.} A closely related line of work aims to distill
multi-step diffusion or flow matching computation in policies into one-step models,
or to train one-step models directly. We focus on the VLA context rather than conventional
diffusion policies~\cite{sheng2025mp1,sochopoulos2025fast,zhang2025flowpolicy,chen2025falcon,jia2024score,ding2025fast,wang2025one,lu2024manicm,gao2026drift,dong2026hybridflow,li2026one,xu2026ada3drift,jia2026action}.
Distillation-based methods~\cite{rdt2, luan2026snapflow, robotics2026habilis}
typically involve two stages: first training a standard multi-step model and then
distilling it into a one-step model, which significantly increases training cost.
Direct one-step training~\cite{chen2026mean, shi2026streamingvla} often requires
model modifications, such as additional inputs in Shortcut Model~\cite{shortcut}
or alternative objectives in MeanFlow~\cite{meanflow}. These changes make it difficult
to fine-tune pretrained VLAs while preserving their strong capabilities. In
contrast, \method requires no architectural modifications and can be seamlessly integrated
into the standard fine-tuning pipeline of flow-based VLAs with the same optimization
objective.

Another related direction explores streaming policies, where the immediate
action is also produced within a single sampling step~\cite{duan2025real,chen2025responsive,hoeg2024streaming}.
However, these approaches update the observation input at every step, which introduces
a substantial inference burden when applied to VLAs. Updating observations
requires a forward pass through the VLM backbone at each step, making real-time VLA
deployment impractical. In contrast, our method is specifically designed for VLAs:
the VLM backbone is forwarded only once per action chunk, thereby avoiding the primary
computational bottleneck in VLA inference. The immediate action is generated
through one-step sampling of the action expert after VLM prefilling, while subsequent
actions are progressively completed through iterations of the action expert based
on the same VLM representations.

\section{Details of Asynchronous Inference Pipeline}
\label{supp:async}

\begin{figure}[t]
  \centering
  \includegraphics[width=\linewidth]{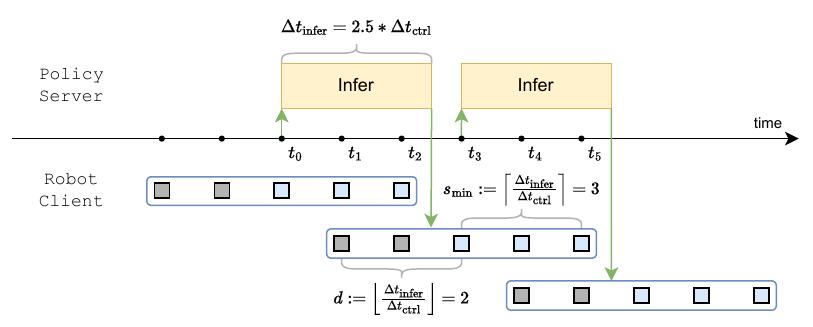}
  \vspace{-15pt}
  \caption{Temporal pipeline of asynchronous inference at a fine-grained level. Suppose
  the inference latency $\tinfer$ is 2.5 times the controller period $\tctrl$,
  resulting in an inference delay of $d=2$ and a minimal execution horizon of
  $s_{\text{min}}=3$.}
  \label{fig:async}
\end{figure}

Here we provide a fine-grained illustration of the asynchronous inference pipeline.
Since the robot controller typically operates at a fixed frequency, all operations
occur with an interval of control period $\tctrl$. As shown in \Cref{fig:async},
the inference latency generally does not align with the controller timesteps in real-world
robotic systems. Therefore, if inference starts at time $t_{0}$, the predicted
actions can only be executed at time
$t_{0}+\lceil \tinfer/\tctrl \rceil \cdot\tctrl$. However, the next action is
expected to be available at $t_{0}+\tctrl$ (one control step later), resulting in
a delay of $d$ actions:
\begin{equation}
  d := \lceil \tinfer/\Delta t_{\text{ctrl}}\rceil-1 = \lfloor \tinfer/\Delta t_{\text{ctrl}}
  \rfloor.
\end{equation}
Note that $\tinfer$ almost never lies exactly on a control-period boundary. Once
the current inference finishes, the next inference can only be triggered at the
subsequent controller timestep (\eg, $t_{3}$ in \Cref{fig:async}). Consequently,
the inference interval is at least $\lceil \tinfer/\tctrl\rceil \cdot \tctrl$.
As discussed in \Cref{tab:react_study}, the inference interval is related to the
execution duration $\texec$, and the minimum feasible execution horizon is
therefore $s_{\text{min}}:= \lceil \tinfer/\Delta t_{\text{ctrl}}\rceil$.

\begin{figure}[t]
  \centering

  \begin{subfigure}
    {0.4\linewidth}
    \centering
    \includegraphics[width=0.9\linewidth]{
      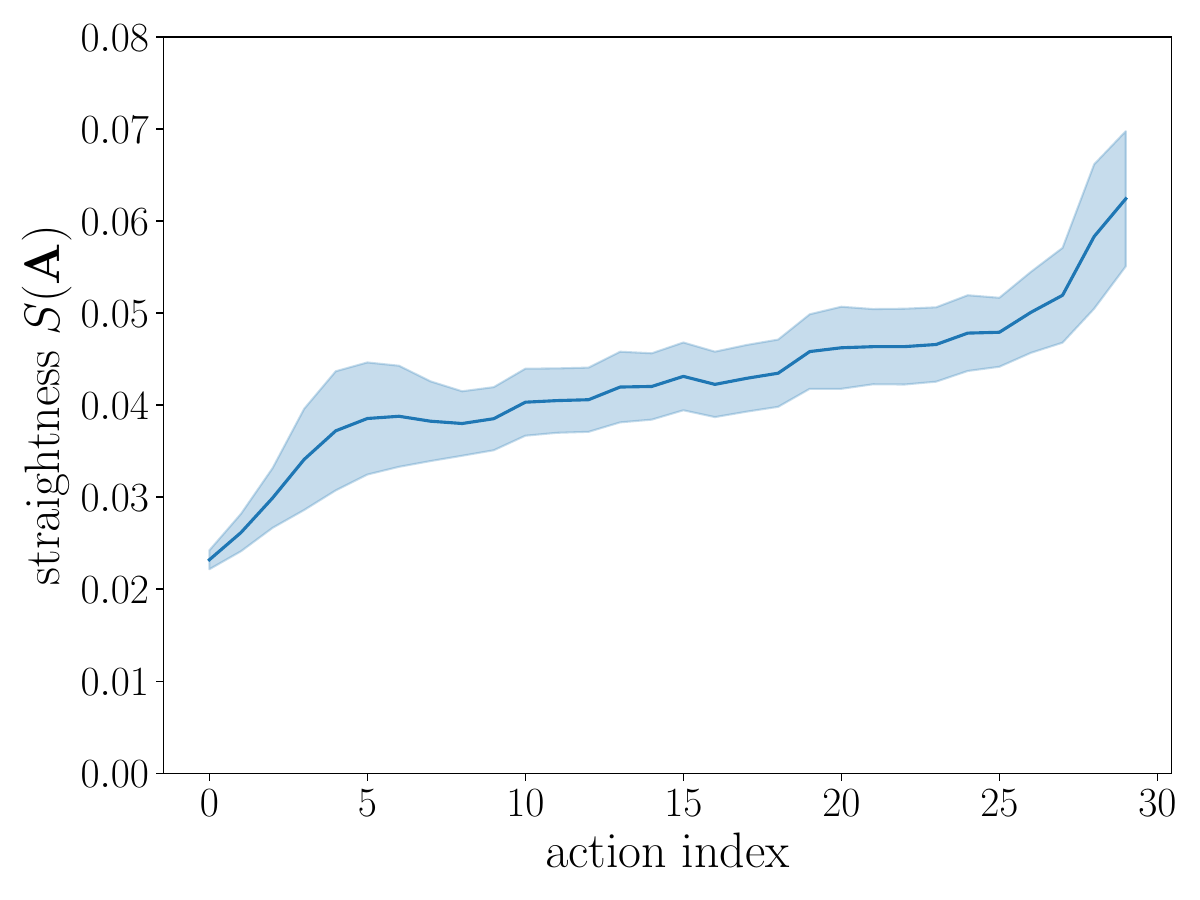
    }
    \vspace{-2mm}
    \caption{}
  \end{subfigure}
  \begin{subfigure}
    {0.4\linewidth}
    \centering
    \includegraphics[width=0.9\linewidth]{
      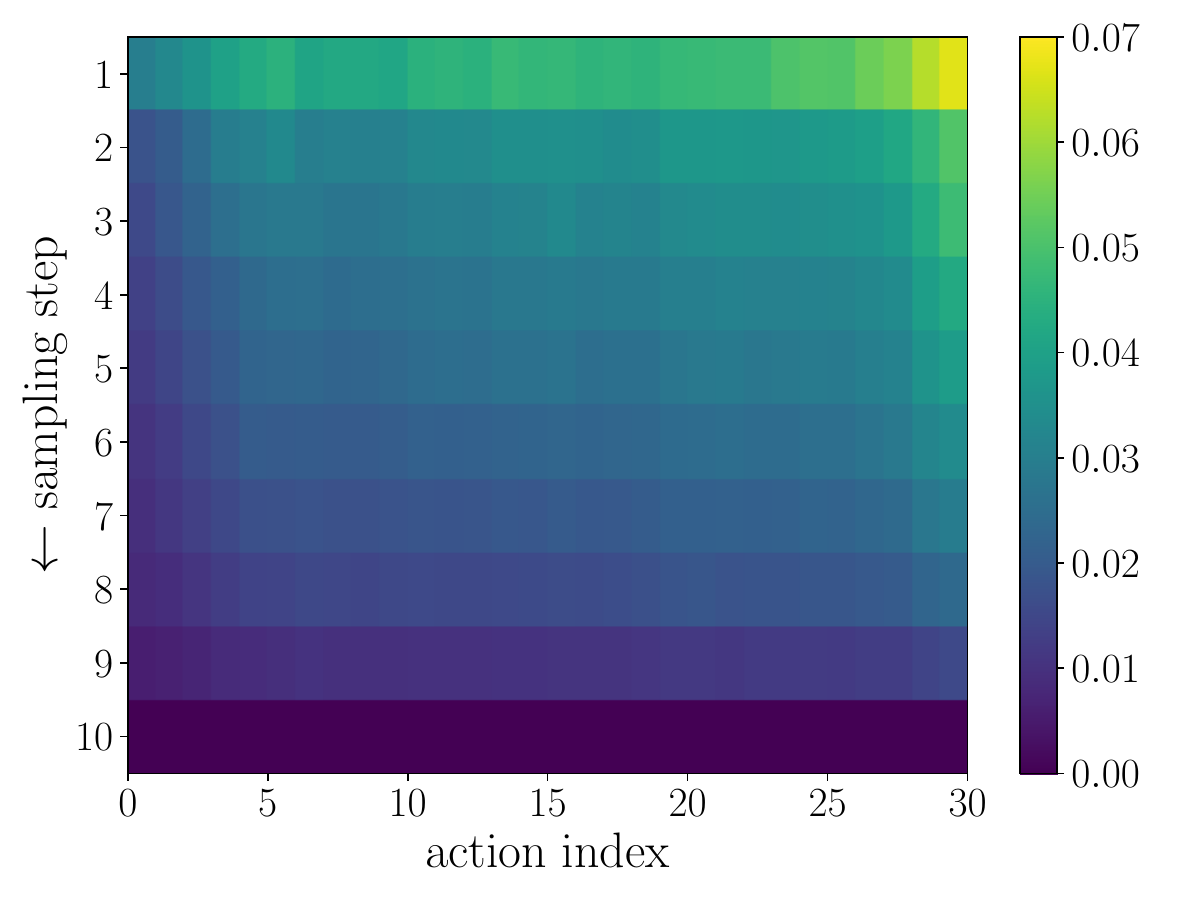
    }
    \vspace{-2mm}
    \caption{}
  \end{subfigure}
  \begin{subfigure}
    {0.4\linewidth}
    \centering
    \includegraphics[width=0.9\linewidth]{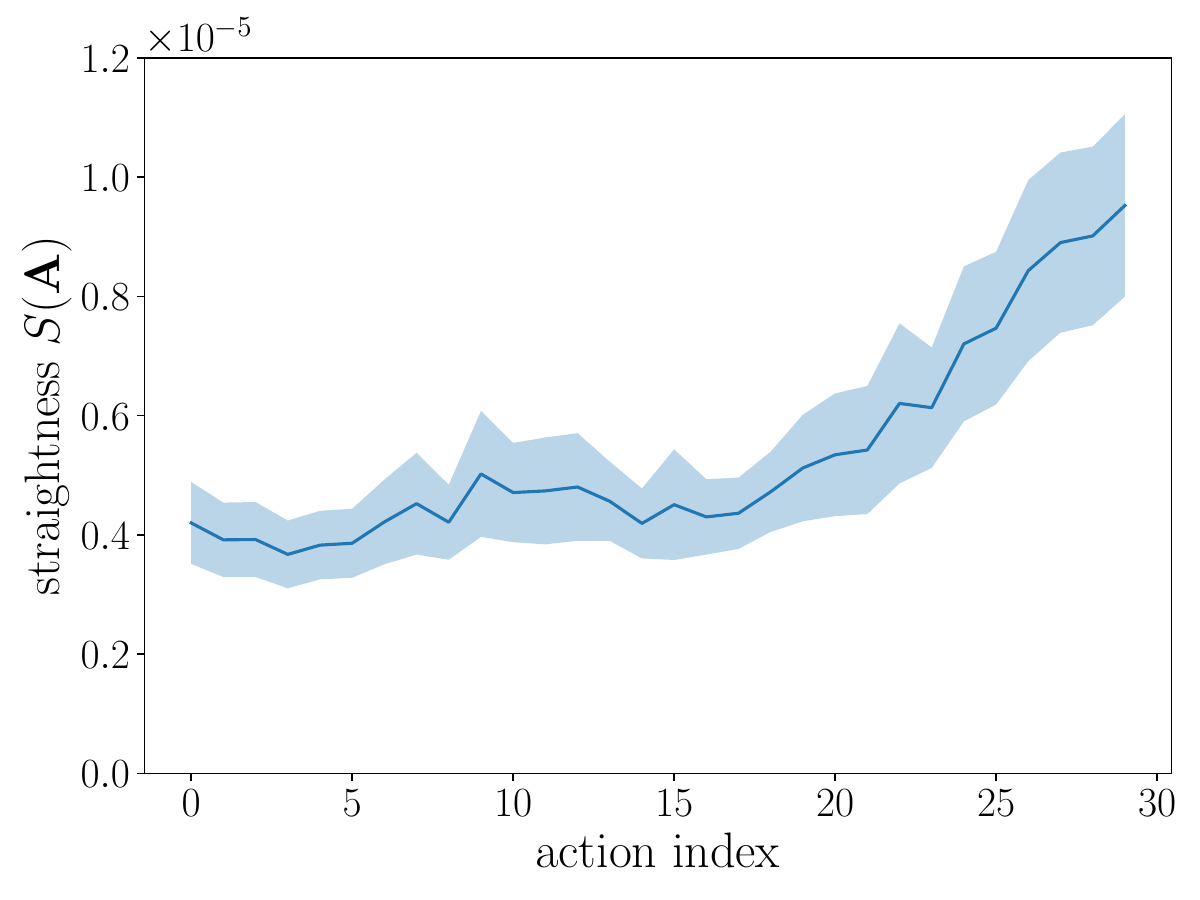}
    \vspace{-2mm}
    \caption{}
  \end{subfigure}
  \begin{subfigure}
    {0.4\linewidth}
    \centering
    \includegraphics[width=0.9\linewidth]{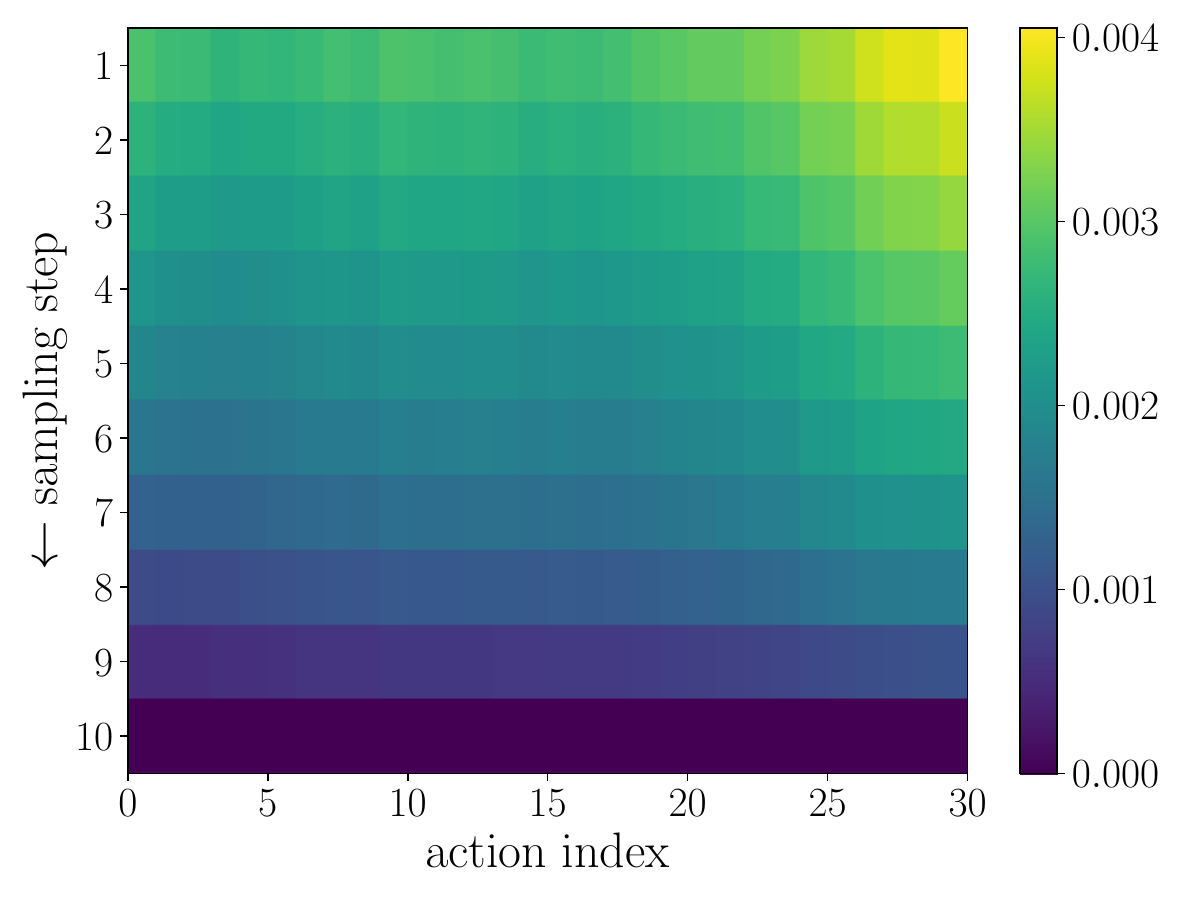}
    \vspace{-2mm}
    \caption{}
  \end{subfigure}
  \begin{subfigure}
    {0.4\linewidth}
    \centering
    \includegraphics[width=0.9\linewidth]{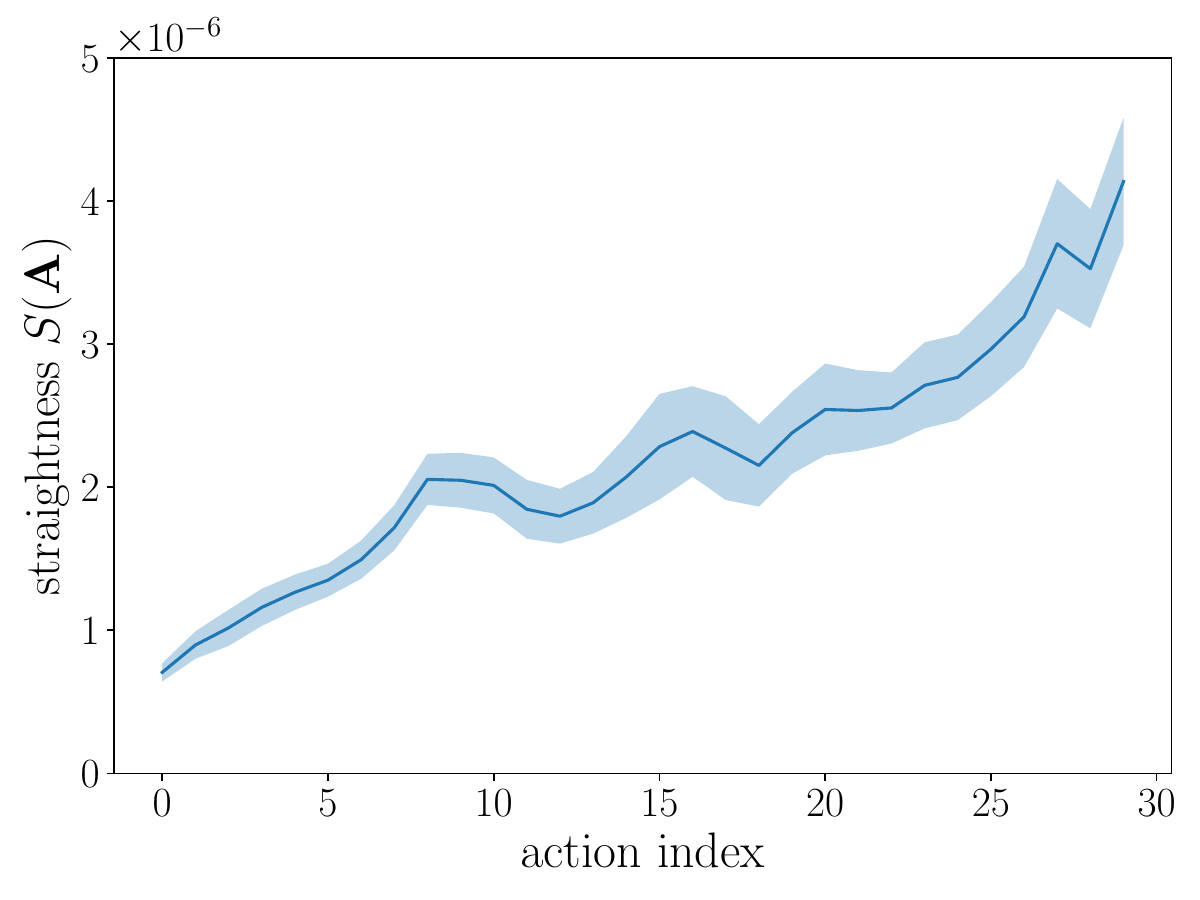}
    \vspace{-2mm}
    \caption{}
  \end{subfigure}
  \begin{subfigure}
    {0.4\linewidth}
    \centering
    \includegraphics[width=0.9\linewidth]{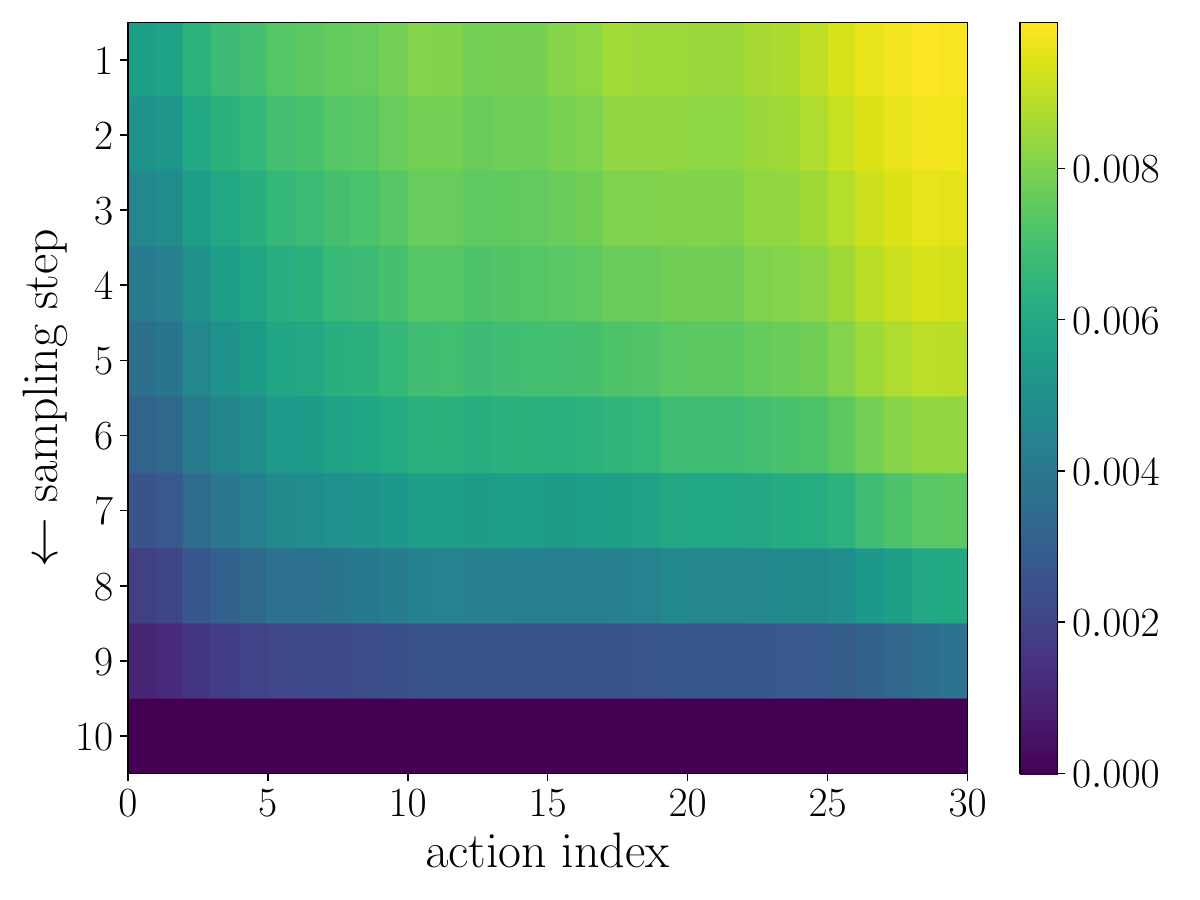}
    \vspace{-2mm}
    \caption{}
  \end{subfigure}
  \caption{Additional visualizations of (a,c,e) straightness $S(\A)$ and (b,d,f)
  differences between the intermediate clean action estimates and the final
  output. Panels (a,b) are computed using a $\pi_{0.5}$ model fine-tuned on Pick
  Beverage task with prediction horizon $H=30$; (c,d) use an X-VLA model fine-tuned
  on the LIBERO simulation dataset with $H=30$; and (e,f) use an X-VLA model fine-tuned
  on the CALVIN ABC simulation dataset with $H=30$. The shaded regions in (a,c,e)
  denote the $\pm2$ SEM across 200 random samples.}
  \label{fig:supp_study}
  \vspace{-10pt}
\end{figure}

\section{Additional Results in Pilot Study}
\label{supp:study}

We present additional results from our pilot study in \Cref{sec:pilot_study}. We
analyze the effect of using a shorter prediction horizon, $H=30$, in \Cref{fig:supp_study}(a,b),
compared with the default setting of $H=50$ in \Cref{fig:pilot_study}(a,b) using
$\pi_{0.5}$. To validate generalization across models and data sources, we
further report results using an X-VLA model on the LIBERO simulation dataset in
\Cref{fig:supp_study}(c,d) and on the CALVIN ABC dataset in \Cref{fig:supp_study}(e,f).
Although the straightness metrics exhibit larger variance under certain conditions,
all metrics reveal a similar non-uniform pattern across action indices, consistent
with \Cref{fig:pilot_study}. These results further support our observation that
earlier actions are easier to sample than later ones, which forms the foundation
for the motivation behind \method.

\section{Additional Methodological Details}
\label{supp:method}

\subsection{Horizon-Aware Schedule with Action Conditioning}
\label{supp:has_ac} We provide details on how the Action Conditioning strategy~\cite{trainingrtc}
is integrated into our Horizon-Aware Schedule. It is also worth noting that action
prefix unimodalizes the distribution for the immediate next action. Because the
subsequent action is heavily constrained by the previous overlapping chunk, it generally
does not demand multi-step expressivity.

In the main paper, we use the superscript $j$ in $\rho^{j}$ and $\tau_{i}^{j}$
to denote the $j$-th sampling step. Here we omit $j$, as the following
description applies to both training and sampling. Specifically, given an action
prefix of length $d < H$, we set the local flow matching timesteps of the prefix
actions to zero\footnote{The notation of timestep is opposite in Training-time
RTC~\cite{trainingrtc}, so they set the timesteps to one in their formulation.}
(\ie, $\tau_{i}\equiv 0$ for $i < d$) and apply an offset $-d$ to the action indices.
The hit time formulation is accordingly updated from \Cref{eq5} to:
\begin{equation}
  \mathbf{u}= \{u_{i}\}= \left( 1 - \left(\frac{i - d}{\max(H - 1 - d, 1)}\right)
  ^{\alpha}\right) \cdot u_{d}, \quad i \in [d+1, H-1], \label{eq:supp2}
\end{equation}
where $u_{d}$ replaces $u_{0}$ as the predefined global timestep at which the \textit{first
valid action} (now the action at index $d$) is finalized. Consequently, the
immediate action still reaches its hit time $u_{d}=(N-1)/N$ with a single
sampling step, ensuring low latency while preserving the temporal consistency of
flow matching. During inference, we simply set $d$ to the real inference delay,
where the action prefix corresponds to the overlapping actions $[ s, s+d-1]$
from the previous chunk.

During training, $d$ is sampled uniformly from 0 to $d_{\text{max}}$ to simulate
varying Time to First Action (TTFA) on distinct devices.
An action prefix mask $\mathbf{m}$ is introduced to mask the loss function:
\begin{equation}
  \mathbf{m}=\{m_{i}\}=\mathbf{1}(i\ge d),\quad i\in[0,H-1], \label{eq:supp3}
\end{equation}
where $\mathbf{1}(\cdot)$ denotes the indicator function. The overall learning
objective is formulated as:
\begin{align}
  \bm{\tau}           & = \{\tau_{i}\} = \begin{cases}0 ,&i < d, \\ \max \left(0, (\rho-u_{i})/(1-u_{i})\right),&i \ge d,\end{cases} \label{eq:supp4}                                                                                                                                         \\
  \mathcal{L}(\theta) & = \mathbb{E}_{\rho\sim\mathcal{U}(0,1),\, d\sim \mathcal{U} \{0,d_{\text{max}}\} }\frac{\left\| \mathbf{m}\odot \left(v_{\theta}(\mathbf{o}_{t}, \A_{t}^{\bm\tau}, \bm{\tau})-(\bm{\epsilon}-\hat{\A}_{t}) \right)\right\|^{2}}{\|\mathbf{m}\|_{1}}, \label{eq:supp5}
\end{align}
where $\odot$ is element-wise multiplication and the $\ell_{1}$ norm $\|\mathbf{m}
\|_{1}$ denotes the number of elements equal to one. Therefore, the loss is computed
only over the suffix actions.

\begin{table}[t!]
  \centering
  \caption{Illustration of inference latency \vs sequential execution. ``Index''
  denotes the index of valid actions in the chunk, excluding delayed actions. ``Time
  Req.'' is the time at which an action is required by the robot controller, while
  ``Time Rec.'' is the time at which an action is received from the policy
  server via the streaming interface, reported as mean $\pm$ standard deviation over
  20 trials. All timestamps are measured relative to the start of inference.}
  \small
  \setlength{\tabcolsep}{8pt}
  \begin{tabular}{c|cccccc}
    \toprule \multirow{2}{*}{GPU}      & \multicolumn{3}{c}{$\pi_{0.5}$} & \multicolumn{3}{c}{X-VLA} \\
                                       & Index                           & Time Req.                & Time Rec.           & Index & Time Req. & Time Rec.           \\
    \midrule \multirow{3}{*}{RTX 4090} & 1                               & 66.7ms                   & 62.1$_{\pm 3.1}$ms  & 1     & 66.7ms    & 44.8$_{\pm 0.3}$ms  \\
                                       & 2                               & 100.0ms                  & 70.2$_{\pm 3.1}$ms  & 2     & 100.0ms   & 52.0$_{\pm 0.1}$ms  \\
                                       & 3                               & 133.3ms                  & 77.1$_{\pm 3.2}$ms  & 3     & 133.3ms   & 59.6$_{\pm 0.3}$ms  \\
    \midrule \multirow{3}{*}{RTX 4060} & 7                               & 266.7ms                  & 238.6$_{\pm 1.9}$ms & 3     & 133.3ms   & 129.2$_{\pm 2.4}$ms \\
                                       & 8                               & 300.0ms                  & 253.3$_{\pm 2.5}$ms & 4     & 166.7ms   & 159.0$_{\pm 3.6}$ms \\
                                       & 9                               & 333.3ms                  & 266.9$_{\pm 2.6}$ms & 5     & 200.0ms   & 186.8$_{\pm 4.3}$ms \\
    \bottomrule
  \end{tabular}
  \label{tab:stream}
\end{table}

\subsection{Streaming Client-Server Interface}
\label{supp:stream}

Unlike conventional action chunking pipelines that transmit an entire action chunk
as a single large payload, the streaming client-server interface in \method
employs a progressive streaming mechanism, decomposing the transmission into a sequence
of high-frequency, smaller packets. While this mechanism increases the total completion
time for a full chunk due to network communication (particularly for the final
action), it aligns naturally with the sequential characteristic of robotic
execution.

Suppose the $i$-th action $\bm{a}_{i}$ is executed at controller time $t$. The system
only requires the $(i+1)$-th action $\bm{a}_{i+1}$ to be available at $t +\tctrl$.
As long as the latency for acquiring $\bm{a}_{i+1}$ remains below the control
period $\tctrl$, the controller avoids pipeline stalls. We provide detailed
illustrations in \Cref{tab:stream}, including latency measurements for two VLAs on
RTX 4090 and RTX 4060 GPUs, respectively. Under this paradigm, the network
latency incurred by subsequent actions is effectively masked by the execution
time of preceding actions. Consequently, the marginal delay for actions at the
end of a chunk becomes functionally negligible and imperceptible during real-time
execution.

\subsection{Algorithms}
The pseudo-code for VLA policy training and inference in \method are provided in
\Cref{alg:train} and \Cref{alg:infer}, respectively.

\begin{algorithm}
  \caption{\method policy training}
  \label{alg:train}
  \begin{algorithmic}
    [1] \Require Training dataset $\mathcal{D}$, pretrained VLA model
    $v_{\theta}$, prediction horizon $H$, HAS parameters $\alpha, u_{d}$, mixing
    probability $p$, max prefix length $d_{\text{max}}$ \Ensure Fine-tuned VLA
    model $v'_{\theta}$ \For{each training step} \State Sample: data
    $(\mathbf{o}_{t}, \hat{\mathbf{A}}_{t}) \sim \mathcal{D}$, noise $\bm{\epsilon}
    \sim\mathcal{N}(\mathbf{0}, \mathbf{I})$, global timestep
    $\rho \sim \mathcal{U}(0,1)$, prefix length $d \sim \mathcal{U}\{ 0, d_{\text{max}}
    \}$, schedule type $z \sim \mathrm{Bernoulli}(p)$ \State Compute action
    prefix mask $\mathbf{m}$:
    \[
      \mathbf{m}=\{m_{i}\}=\mathbf{1}(i\ge d),\quad i\in[0,H-1],
    \]
    \If{$z = 1$}\Comment{Use Horizon-Aware Schedule} \State Compute hit times
    $\mathbf{u}$:
    \[
      \mathbf{u}= \{u_{i}\}= \left( 1 - \left(\frac{i - d}{\max(H - 1 - d, 1)}\right
      )^{\alpha}\right) \cdot u_{d}, \quad i \in [d+1, H-1],
    \]
    \State Compute local timesteps $\bm\tau$:
    \[
      \bm{\tau}= \{\tau_{i}\} =
      \begin{cases}
        0 ,                                          & i < d,   \\
        \max \left(0, (\rho-u_{i})/(1-u_{i})\right), & i \ge d,
      \end{cases}
    \]
    \Else \Comment{Use Constant Schedule with Action Conditioning} \State Set local
    timesteps $\bm\tau$: \\
    \[
      \bm{\tau}= \{\tau_{i}\} =
      \begin{cases}
        0 ,   & i < d,   \\
        \rho, & i \ge d,
      \end{cases}
    \]
    \EndIf \State Construct noisy actions:
    \[
      \mathbf{A}_{t}^{\bm{\tau}}= \bm{\tau}\odot \bm{\epsilon}+ (1-\bm{\tau})\odot
      \hat{\mathbf{A}}_{t}
    \]
    \State Compute masked loss:
    \[
      \mathcal{L}(\theta) = \frac{\left\| \mathbf{m}\odot \left(v_{\theta}(\mathbf{o}_{t},
      \A_{t}^{\bm\tau}, \bm{\tau})-(\bm{\epsilon}-\hat{\A}_{t}) \right)\right\|^{2}}{\|\mathbf{m}\|_{1}}
      ,
    \]
    \State Update model parameters $\theta$ using
    $\nabla_{\theta}\mathcal{L}(\theta)$ \EndFor
  \end{algorithmic}
\end{algorithm}

\begin{algorithm}
  [t!]
  \caption{\method policy inference}
  \label{alg:infer}
  \begin{algorithmic}
    [1] \Require Fine-tuned VLA model $v'_{\theta}$, sampling steps $N$, prediction
    horizon $H$, HAS parameters $\alpha, u_{d}$, observation $\mathbf{o}_{t}$,
    delay $d$, action prefix $\mathbf{A}^{\text{pre}}$, execution horizon $s$
    \State Initialize $\mathbf{A}_{t}\sim \mathcal{N}(\mathbf{0}, \mathbf{I})$
    \State Compute hit times $\mathbf{u}$
    \[
      \mathbf{u}= \{u_{i}\}= \left( 1 - \left(\frac{i - d}{\max(H - 1 - d, 1)}\right
      )^{\alpha}\right) \cdot u_{d}, \quad i \in [d+1, H-1],
    \]
    \State Forward VLM backbone with $\mathbf{o}_{t}$ \For{$j=1$ to $N$} \State Set
    global timestep $\rho^{j}= (N-j+1)/N$ \State Compute local timesteps $\bm\tau
    ^{j}$ w.r.t. $\rho^{j}$:
    \[
      \bm{\tau}^{j}= \{\tau_{i}^{j}\} =
      \begin{cases}
        0 ,                                              & i < d,   \\
        \max \left(0, (\rho^{j}-u_{i})/(1-u_{i})\right), & i \ge d,
      \end{cases}
    \]
    \State Similarly, compute next local timesteps $\bm\tau^{j+1}$ w.r.t. $\rho^{j+1}$
    \Comment{$\rho^{N+1}:=0$} \State Overwrite actions prefix
    $\mathbf{A}_{t}[0:d-1]$ with $\mathbf{A}^{\mathrm{pre}}$ \State Predict
    velocity
    $\mathbf{v}^{j}= v_{\theta}(\mathbf{o}_{t}, \mathbf{A}_{t}, \bm{\tau}^{j})$\Comment{Forward AE only}
    \State Compute delta timestep
    $\Delta \bm{\tau}=\bm{\tau}^{j+1}-\bm{\tau}^{j}$ \State Euler update
    $\mathbf{A}_{t}\gets \mathbf{A}_{t}+ \mathbf{v}^{j}\odot \Delta \bm{\tau}$
    \For{each valid action index $i \in [d, H-1]$} \If{$\tau_{i}^{j+1}= 0$ \textbf{and} action $i$ has not been streamed}
    \State Dispatch $\bm{a}_{t+i}$ to client immediately \Comment{Streaming Interface}
    \EndIf \EndFor \If{all actions within execution horizon $[d, d+s-1]$ have been finalized}
    \State \textbf{break} \Comment{Early stopping} \EndIf \EndFor
  \end{algorithmic}
\end{algorithm}

\clearpage
\newpage
\section{Implementation Details}
\label{supp:imple}

\subsection{Real-world Experiments}
\label{supp:real}

\begin{figure}
  \centering
  \includegraphics[width=0.65\linewidth]{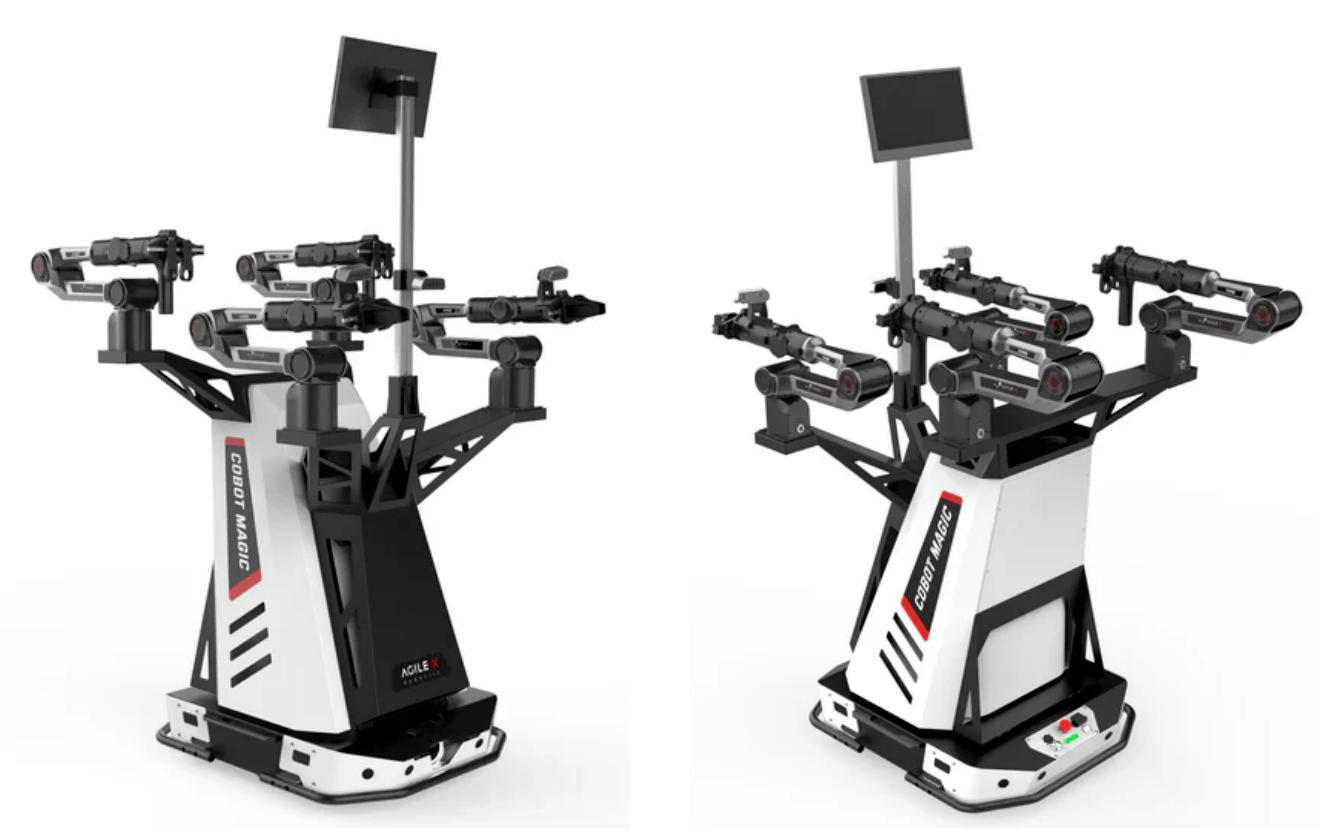}
  \caption{AgileX Cobot Magic robotic platform with Piper arms.}
  \label{fig:cobot}
\end{figure}

\noindent
\textbf{Hardware Setup.} We use the AgileX Cobot Magic robotic platform shown in
\Cref{fig:cobot}, designed in the Mobile Aloha style~\cite{act,mobilealoha}. It
is equipped with four 6-DoF Piper robot arms: two leader arms for human teleoperation
and two follower arms for data collection and policy inference. A multi-camera
setup is used, consisting of one front camera (RealSense D455) and two wrist cameras
(RealSense D435), each mounted on a follower arm.

\begin{figure}
  \centering
  \includegraphics[width=\linewidth]{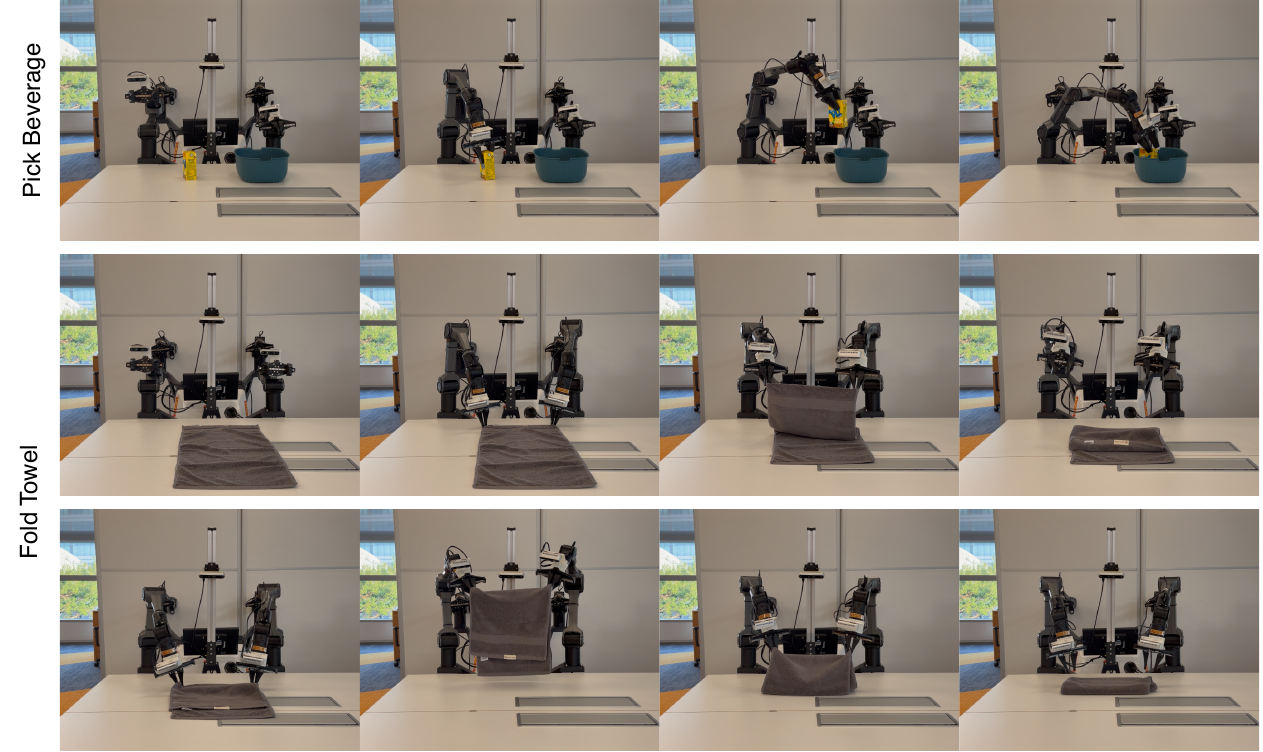}
  \caption{Visualization of Pick Beverage and Fold Towel tasks.}
  \label{fig:add_tasks}
\end{figure}

\noindent
\textbf{Tasks.} We evaluate three real-robot tasks: ``Table Tennis'', ``Pick Beverage'',
and ``Fold Towel''. The visualization of Table Tennis task is already provided in
the main paper, while illustrations of the other two tasks are shown in
\Cref{fig:add_tasks}. We collect 335 demonstration episodes (approximately 14
minutes) for Table Tennis task and 150 episodes for the other two tasks via human
teleoperation. All data is recorded at 30 FPS.

The language instructions for the tasks are as follows:
\begin{itemize}[noitemsep, topsep=0pt,leftmargin=*]
  \item Table Tennis: ``Hit the table tennis ball to the opponent.''

  \item Pick Beverage: ``Pick up the beverage and put it in the plastic basket.''

  \item Fold Towel: ``Fold the towel.''
\end{itemize}

\noindent
\textbf{Evaluation.} The tasks are evaluated with a fixed number of rollouts: 15
trials for Table Tennis, 35 trials for Pick Beverage, and 10 trials for Fold
Towel. For Pick Beverage and Fold Towel, we follow predefined test cases with
fixed object types, positions, and orientations to ensure fair comparison. For Table
Tennis, we regulate the ball speed to the best of our ability.

We define fine-grained evaluation metrics to more precisely assess the real-world
performance of the models under a limited number of rollouts. Results are reported
by averaging scores across sub-steps and rollouts. The scoring criteria are as
follows:

\begin{enumerate}[noitemsep, topsep=0pt, leftmargin=*]
  \item Table Tennis
    \begin{itemize}
      \item Step 1: Hitting the table tennis ball with the racket
        \begin{itemize}
          \item 0 point: The robot misses the ball.

          \item 0.5 point: The robot returns the ball but produces a weak hit
            due to reaction latency; the ball travels only a short distance before
            landing on the table. The distance is measured using a marked line
            on the table.

          \item 1 point: The robot performs a powerful return, and the ball travels
            a significant distance before landing on the table.
        \end{itemize}
    \end{itemize}

  \item Pick Beverage
    \begin{itemize}
      \item Step 1: Grasping the beverage
        \begin{itemize}
          \item 0 point: The robot fails to grasp the object.

          \item 0.5 point: The robot grasps the object after multiple (up to five)
            attempts and recovers from errors.

          \item 1 point: The robot successfully grasps the object on the first
            attempt.
        \end{itemize}

      \item Step 2: Placing the beverage into the basket
        \begin{itemize}
          \item 0 point: The object is not placed into the basket or is dropped.

          \item 0.5 point: The robot places the object into the basket but
            causes a collision.

          \item 1 point: The robot successfully places the object into the
            basket on the first attempt.
        \end{itemize}
    \end{itemize}

  \item Fold Towel
    \begin{itemize}
      \item Step 1: Grasping the towel
        \begin{itemize}
          \item 0 point: The robot fails to grasp both sides of the towel.

          \item 0.5 point: The robot grasps both sides after multiple (up to five)
            attempts and recovers from errors.

          \item 1 point: The robot successfully grasps both sides on the first
            attempt.
        \end{itemize}

      \item Step 2: Forward folding the towel
        \begin{itemize}
          \item 0 point: The robot fails to fold the towel forward (\eg,
            slippage from the grippers).

          \item 0.5 point: The towel is folded forward but not well aligned.

          \item 1 point: The robot folds the towel forward perfectly.
        \end{itemize}

      \item Step 3: Grasping the towel again
        \begin{itemize}
          \item 0 point: The robot fails to grasp both sides of the towel.

          \item 0.5 point: The robot grasps both sides after multiple (up to five)
            attempts and recovers from errors.

          \item 1 point: The robot successfully grasps both sides on the first
            attempt.
        \end{itemize}

      \item Step 4: Backward folding the towel
        \begin{itemize}
          \item 0 point: The robot fails to fold the towel backward (\eg,
            slippage from the grippers).

          \item 0.5 point: The towel is folded backward but not well aligned.

          \item 1 point: The robot folds the towel backward perfectly.
        \end{itemize}
    \end{itemize}
\end{enumerate}

\begin{table}[t]
  \centering
  \caption{Hyperparameter settings for fine-tuning VLA models.}
  \small
  \setlength{\tabcolsep}{5pt}
  \renewcommand{\arraystretch}{1.2}
  \begin{tabular}{c|cccc}
    \toprule Model                  & \multicolumn{2}{c}{$\pi_{0.5}$} & \multicolumn{2}{c}{X-VLA} \\
    Task                            & AgileX                          & Simulation               & AgileX      & Simulation  \\
    \midrule Prediction Horizon $H$ & 50                              & 10                       & 30          & 30          \\
    Action Space                    & Relative Joint                  & Delta EEF                & Abs EE6D    & Abs EE6D    \\
    Global Batch Size               & 128                             & 256                      & 128         & 128         \\
    Training Steps                  & 50k                             & 30k                      & 50k         & 60k         \\
    Optimizer                       & AdamW                           & AdamW                    & AdamW       & AdamW       \\
    Weight Decay                    & 0                               & 0                        & 0           & 0           \\
    Betas                           & (0.9, 0.95)                     & (0.9, 0.95)              & (0.9, 0.95) & (0.9, 0.95) \\
    Base LR                         & 2.5e-5                          & 5e-5                     & 1e-4        & 1e-4        \\
    LR Scheduler                    & Cosine Decay                    & Cosine Decay             & Constant    & Constant    \\
    Warmup Steps                    & 1k                              & 10k                      & 2k          & 2k          \\
    Grad Norm Clip                  & 1.0                             & 1.0                      & 1.0         & 1.0         \\
    EMA Decay                       & 0.99                            & 0.999                    & N/A         & N/A         \\
    \bottomrule
  \end{tabular}
  \label{tab:hyperparam}
\end{table}

\begin{table}
  \centering
  \caption{Settings of inference delay $d$ and execution horizon $s$ on RTX 4090
  and RTX 4060 GPUs. ``Async'' refers to Naive Async and Training-time RTC.}
  \small
  \setlength{\tabcolsep}{7pt}
  \renewcommand{\arraystretch}{1.2}
  \begin{tabular}{cc|cccc}
    \toprule \multirow{2}{*}{Task}                      & \multirow{2}{*}{Method} & \multicolumn{2}{c}{RTX 4090} & \multicolumn{2}{c}{RTX 4060} \\
                                                        &                         & $d$                          & $s$                         & $d$ & $s$ \\
    \midrule \multirow{3}{*}{Table Tennis}              & Sync                    & 4                            & 5                           & 10  & 11  \\
                                                        & Async                   & 4                            & 5                           & 10  & 11  \\
                                                        & \method                 & 3                            & 4                           & 8   & 10  \\
    \midrule \multirow{3}{*}{Pick Beverage, Fold Towel} & Sync                    & 4                            & 50                          & 10  & 50  \\
                                                        & Async                   & 4                            & 46                          & 10  & 40  \\
                                                        & \method                 & 3                            & 47                          & 8   & 42  \\
    \bottomrule
  \end{tabular}
  \label{tab:client_setup}
\end{table}

\subsection{Simulation Benchmarks}

\noindent
\textbf{LIBERO.} The LIBERO benchmark~\cite{libero} consists of four task suites:
Spatial, Object, Goal, and 10 (Long), each targeting distinct aspects of
embodied capabilities. Following prior work~\cite{pi05, xvla}, we train a single
policy across all four suites. For training $\pi_{0.5}$, we use the dataset
provided by OpenVLA\footnote{HuggingFace: \texttt{openvla/modified\_libero\_rlds}}~\cite{openvla}
and convert it to the LeRobot format~\cite{lerobot} using the script in \texttt{openpi}.
While for training X-VLA, we use the HDF5-format dataset provided by X-VLA
authors\footnote{HuggingFace: \texttt{2toINF/Libero-XVLA-format}}, which aligns
with the EE6D action space. Each suite contains 10 tasks, and every task is
evaluated over 50 trials.

\noindent
\textbf{CALVIN.} The CALVIN benchmark~\cite{calvin} comprises 34 tasks with unconstrained
language instructions spanning diverse manipulation skills. We adopt the widely
used ABC$\rightarrow$D evaluation setting and use the LeRobot-format dataset\footnote{HuggingFace:
\texttt{InternRobotics/InternData-Calvin\_ABC}}. Each model is evaluated on 1,000
unique instruction chains, where each chain consists of five consecutive tasks.
Performance is measured by the average number of successfully completed tasks
per chain.

\noindent
\textbf{Kinetix.} The Kinetix benchmark\footnote{GitHub: \texttt{Physical-Intelligence/real-time-chunking-kinetix}}
is proposed by RTC~\cite{rtc} and contains 12 dynamic tasks in the Kinetix
simulation environment~\cite{kinetix}. Since these tasks involve dynamic motions
with force-based control, they are suitable for evaluating asynchronous
execution methods under varying inference delays. However, the policy used in
this benchmark is a simple 4-layer MLP rather than a VLA model, so it is not the
primary focus of our study.

Following prior work~\cite{rtc, trainingrtc}, the policy uses a prediction
horizon of $H=8$. During training, the simulated delay is randomly sampled from
0 to 4 with exponentially decreasing weights. We resume training from the 24th epoch
and fine-tune the policy for 8 epochs using our mixed schedule. The number of
sampling steps is set to 5, and accordingly we set $u_{d}=0.8$. Performance is
evaluated over 2,048 rollouts, and we report the average success rates.

\begin{table}[t]
  \centering
  \caption{Detailed comparison of reaction capability on RTX 4090 and RTX 4060
  GPUs. The inference latency $\tinfer$ corresponds to TTFA, and execution
  duration $\texec$ is computed as $s_{\text{min}}\cdot\tctrl$. The distribution
  of reaction time $\dreact$ is $\mathcal{U}(\tinfer, 2*\tinfer+\texec)$ for Sync,
  and $\mathcal{U}(\tinfer, \tinfer+\texec)$ for Async and \method.}
  \small
  \setlength{\tabcolsep}{5pt}
  \renewcommand{\arraystretch}{1.2}
  \begin{tabular}{cc|cccccccc}
    \toprule \multirow{2}{*}{Model}       & \multirow{2}{*}{Method} & \multicolumn{3}{c}{RTX 4090} & \multicolumn{3}{c}{RTX 4060} \\
                                          &                         & $\tinfer$                    & $\texec$                    & $\dreact$                   & $\tinfer$        & $\texec$         & $\dreact$                    \\
    \midrule \multirow{3}{*}{$\pi_{0.5}$} & Sync                    & 80.0ms                       & 100.0ms                     & $\mathcal{U}(80.0, 260.0)$  & 303.3ms          & 333.3ms          & $\mathcal{U}(303.3,939.9)$   \\
                                          & Async                   & 80.0ms                       & 100.0ms                     & $\mathcal{U}(80.0, 180.0)$  & 303.3ms          & 333.3ms          & $\mathcal{U}(303.3, 636.6)$  \\
                                          & \method                 & \textbf{62.1}ms              & 100.0ms                     & $\mathcal{U}(62.1, 162.1)$  & \textbf{238.6}ms & \textbf{266.7}ms & $\mathcal{U}(238.6, 505.3)$  \\
    \midrule \multirow{3}{*}{X-VLA}       & Sync                    & 113.7ms                      & 133.3ms                     & $\mathcal{U}(113.7, 360.7)$ & 399.5ms          & 400.0ms          & $\mathcal{U}(399.5, 1199.0)$ \\
                                          & Async                   & 113.7ms                      & 133.3ms                     & $\mathcal{U}(113.7, 247.0)$ & 399.5ms          & 400.0ms          & $\mathcal{U}(399.5, 799.5)$  \\
                                          & \method                 & \textbf{44.8}ms              & \textbf{66.7}ms             & $\mathcal{U}(44.8, 111.5)$  & \textbf{129.2}ms & \textbf{200.0}ms & $\mathcal{U}(129.2, 329.2)$  \\
    \bottomrule
  \end{tabular}
  \label{tab:react_supp}
\end{table}

\subsection{Training Setup}
We use the official codebases of $\pi_{0.5}$\footnote{Github: \texttt{Physical-Intelligence/openpi}}
and X-VLA\footnote{Github: \texttt{2toinf/X-VLA}} to fine-tune the VLA models. We
follow their default or recommended configurations, with key hyperparameters
listed in \Cref{tab:hyperparam}. Our real-robot dataset is recorded in the absolute
joint space. Actions are converted to the relative joint space by computing
offsets from the proprioceptive states, and to the absolute EE6D space using forward
kinematics. All models are trained on 8 NVIDIA A800 GPUs with 80GB VRAM from
cloud provider, requiring 1-2 days for $\pi_{0.5}$ and 6-7 hours for X-VLA.

Regarding the hyperparameters of our proposed Horizon-Aware Schedule, we use
$\alpha =0.6$ to control the hit times and $p=0.5$ for the mixed schedule by
default. When training X-VLA on LIBERO and CALVIN, we set $\alpha$ to 0.7; ablation
studies are provided in \Cref{supp:abla}. Since both VLAs use 10 sampling steps
for flow matching, we set $u_{0}=0.9$ to ensure that the immediate action requires
only a single step, as described in the main paper.

Although the prediction horizon is 50 or 30, we set the maximum prefix length $d_{\text{max}}$
to 10 during fine-tuning real-world tasks, which simulate TTFA up to 333.3ms.
This design is motivated by practical considerations: inference delays exceeding
333.3ms can render real-time reactive control functionally infeasible for
certain tasks. By limiting the maximum prefix length, we prevent the model from
wasting capacity on physically unrealistic scenarios. This range already provides
sufficient coverage even for an RTX 4060 GPU, thereby ensuring better sample
efficiency and smoother flow matching in the critical low-latency regime.

\subsection{Robot Deployment Setup}
As mentioned in the main paper, the ROS system on the robot runs at a fixed 30Hz
frequency, \ie, with a control period $\tctrl = 33.3$ms. The robot is connected to
policy server via LAN and communicates through websocket protocol. We define the
inference delay as $d := \lfloor \tinfer / \Delta t_{\text{ctrl}}\rfloor$ and
the minimum execution horizon as
$s_{\text{min}}:= \lceil \tinfer / \Delta t_{\text{ctrl}}\rceil$. To account for
additional overhead from network communication, local processing, memory I/O,
and other system costs, we set these values slightly larger in real-robot deployment,
as listed in \Cref{tab:client_setup}.

For the table tennis task, we set the execution horizon to $s_{\text{min}}$ to achieve
optimal reaction capability. For other tasks that do not heavily depend on high-frequency
control, we set the execution horizon to the number of valid actions ($H$ for
Sync, $H-d$ for Async and \method) to ensure smooth action trajectories.

\begin{figure}[t]
  \centering
  \includegraphics[width=\linewidth]{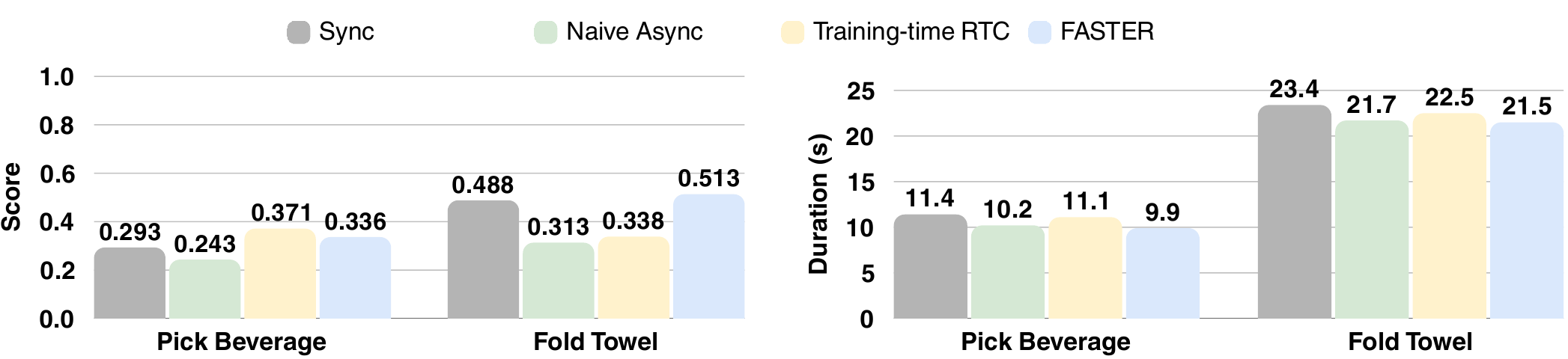}
  \vspace{-3mm}
  \caption{Comparison of real-world performance and task completion duration on
  two real-world tasks using X-VLA. Note that the duration is computed only from
  successful rollouts, and therefore is not directly comparable to the results
  in \Cref{fig:exp_tasks}.}
  \label{fig:tasks_xvla}
\end{figure}

\begin{table}[t]
  \centering
  \caption{Performance statistics for the Table Tennis task. The 96\% confidence
  intervals are computed using bootstrap resampling.}
  \small
  \setlength{\tabcolsep}{10pt}
  \begin{tabular}{cc|cccc}
    \toprule \multirow{2}{*}{Model}       & \multirow{2}{*}{Method} & \multicolumn{2}{c}{RTX 4090} & \multicolumn{2}{c}{RTX 4060} \\
                                          &                         & Mean                         & 96\% CI                     & Mean  & 96\% CI        \\
    \midrule \multirow{4}{*}{$\pi_{0.5}$} & Sync                    & 0.000                        & (0.000, 0.000)              & 0.000 & (0.000, 0.000) \\
                                          & Naive Async             & 0.200                        & (0.075, 0.325)              & 0.200 & (0.067, 0.333) \\
                                          & Training-time RTC       & 0.533                        & (0.300, 0.767)              & 0.300 & (0.133, 0.500) \\
                                          & \method                 & 0.800                        & (0.667, 0.933)              & 0.467 & (0.333, 0.567) \\
    \bottomrule
  \end{tabular}
  \label{tab:tt_ci}
\end{table}

\begin{table}[t]
  \centering
  \caption{Performance statistics for the Pick Beverage and Fold Towel task. The
  96\% confidence intervals are computed using bootstrap resampling.}
  \small
  \setlength{\tabcolsep}{5pt}
  \begin{tabular}{cc|cccc|cccc}
    \toprule \multirow{3}{*}{Model}       & \multirow{3}{*}{Method} & \multicolumn{4}{c|}{Pick Beverage} & \multicolumn{4}{c}{Fold Towel} \\
                                          &                         & \multicolumn{2}{c}{Score}          & \multicolumn{2}{c|}{Duration} & \multicolumn{2}{c}{Score} & \multicolumn{2}{c}{Duration} \\
                                          &                         & Mean                               & 96\% CI                       & Mean                      & Std                         & Mean  & 96\% CI        & Mean & Std \\
    \midrule \multirow{4}{*}{$\pi_{0.5}$} & Sync                    & 0.879                              & (0.786, 0.950)                & 13.0                      & 0.9                         & 0.788 & (0.600, 0.925) & 24.7 & 0.5 \\
                                          & Naive Async             & 0.957                              & (0.886, 1.000)                & 12.5                      & 1.2                         & 0.825 & (0.613, 0.988) & 24.0 & 2.4 \\
                                          & Training-time RTC       & 0.950                              & (0.879, 0.993)                & 11.9                      & 2.4                         & 0.888 & (0.700, 1.000) & 20.7 & 0.4 \\
                                          & \method                 & 0.957                              & (0.886, 1.000)                & 12.0                      & 0.9                         & 0.963 & (0.925, 1.000) & 20.5 & 0.4 \\
    \bottomrule
  \end{tabular}
  \label{tab:bf_ci}
\end{table}

\section{Additional Experimental Results}
\label{supp:exp}

\subsection{Additional Analysis on Reaction Speed}

We provide \Cref{tab:react_supp} to complement the reaction time details in \Cref{tab:react}.
The probabilities reported in \Cref{tab:react_prob} are also derived from the
distributions $\dreact$, defined as the probability that one independently sampled
$\treact$ from a method is smaller than the other method. As discussed in the
main paper, \method is deterministically superior on X-VLA, since its upper
bound of reaction time is lower than the baselines' lower bound: 111.5ms \vs 113.7ms
on RTX 4090, and 329.2ms \vs 399.5ms on RTX 4060.

\subsection{Additional Real-world Experiments}
We evaluate \method on real-world Pick Beverage and Fold Towel tasks using the X-VLA~\cite{xvla}
model and present the results in \Cref{fig:tasks_xvla}. We observe that though X-VLA
performs significantly worse than the state-of-the-art VLA $\pi_{0.5}$, our
method consistently achieves better or comparable results in terms of both
completion scores and task duration. It is worth noting that Naive Async often fails
due to unstable actions caused by inter-chunk discontinuities.

We supplement the performance statistics of \Cref{fig:pp} and \Cref{fig:exp_tasks}
in \Cref{tab:tt_ci} and \Cref{tab:bf_ci}, respectively.

\subsection{Simulation Benchmarks}
\label{supp:sim}

We conduct experiments on two widely used simulation benchmarks in VLA research:
LIBERO~\cite{libero} and CALVIN~\cite{calvin}. Although these simulation environments
are not directly affected by inference latency, they provide a controlled
protocol for evaluating whether \method preserves the original model performance.
As shown in \Cref{tab:sim}, HAS maintains competitive performance on both
benchmarks, with limited degradation despite its aggressive action sampling strategy.
These results suggest that, although FASTER may slightly affect long-horizon accuracy
in simulation, it provides a favorable trade-off for latency-sensitive real-world
settings.

\begin{table}[t]
  \centering
  \caption{Comparison of performance on simulation benchmarks LIBERO and CALVIN.
  ``$*$'' indicates experimental results evaluated on our cluster with official checkpoints.}
  \small
  \setlength{\tabcolsep}{4.5pt}
  \begin{tabular}{c|ccccc|cccccc}
    \toprule \multirow{2}{*}{Method} & \multicolumn{5}{c|}{LIBERO} & \multicolumn{6}{c}{CALVIN ABC$\rightarrow$D} \\
                                     & Spatial                     & Object                                      & Goal & 10   & Avg. & 1    & 2    & 3    & 4    & 5    & Avg. Len \\
    \midrule                          %
    $\pi_{0.5}$                      & 98.8                        & 98.2                                        & 98.0 & 92.4 & 96.9 & 94.2 & 88.7 & 85.7 & 83.2 & 79.5 & 4.313    \\
    $\pi_{0.5}$+\method              & 98.6                        & 97.8                                        & 97.8 & 91.6 & 96.5 & 95.1 & 89.1 & 85.0 & 81.9 & 78.1 & 4.292    \\
    \midrule X-VLA$^{*}$             & 97.8                        & 99.4                                        & 97.8 & 96.8 & 98.0 & 95.7 & 89.8 & 82.4 & 77.0 & 70.2 & 4.151    \\
    X-VLA+\method                    & 97.8                        & 99.4                                        & 98.2 & 93.0 & 97.1 & 98.1 & 91.4 & 82.4 & 72.9 & 64.5 & 4.093    \\
    \bottomrule
  \end{tabular}
  \label{tab:sim}
  \vspace{-5mm}
\end{table}

We further evaluate on the Kinetix benchmark~\cite{rtc} to compare with state-of-the-art
real-time methods. We include additional baselines: Bidirectional Decoding (BID)~\cite{bid},
Inference-time RTC~\cite{rtc}, VLASH~\cite{vlash}, and REMAC~\cite{remac}. Since
flow-matching sampling is applied to the entire model in this benchmark, \method
reduces the inference latency of the immediate action by $5\times$. We conduct fair
comparisons under the same wall-clock inference budget: given the maximum
supported delay is 4, we compare the baselines at $d=4$ with \method at $d=1$ under
the same execution horizon. As shown in \Cref{tab:kinetix}, \method outperforms all
baselines by a clear margin, highlighting the importance of reducing reaction
latency in asynchronous execution, even in simulated environments.

\begin{table}[t]
  \centering
  \caption{Comparison of performance on Kinetix benchmark under the same
  inference budget.}
  \small
  \setlength{\tabcolsep}{10pt}
  \begin{tabular}{c|cccc}
    \toprule Method                      & $d$ & $s$ & Solve Rate     \\
    \midrule Naive Async                 & 4   & 4   & 0.492          \\
    BID~\cite{bid}                       & 4   & 4   & 0.553          \\
    Inference-time RTC~\cite{rtc}        & 4   & 4   & 0.614          \\
    Training-time RTC~\cite{trainingrtc} & 4   & 4   & 0.726          \\
    REMAC~\cite{remac}                   & 4   & 4   & 0.779          \\
    VLASH~\cite{vlash}                   & 4   & 4   & 0.813          \\
    \method                              & 1   & 4   & \textbf{0.869} \\
    \bottomrule
  \end{tabular}
  \label{tab:kinetix}
\end{table}

\subsection{Ablation Study}
\label{supp:abla}

We conduct ablation studies to analyze the impact of our proposed methods on task
performance. Since real-world experiments are difficult to repeat with a large
number of rollouts and are easily affected by environmental factors, we use simulation
benchmarks for more controlled evaluation. Moreover, as the LIBERO benchmark is highly
saturated, we focus on CALVIN to better highlight long-horizon performance.

\begin{figure}[t]
  \centering
  \begin{minipage}{0.42\linewidth}
    \centering
    \includegraphics[width=0.9\linewidth]{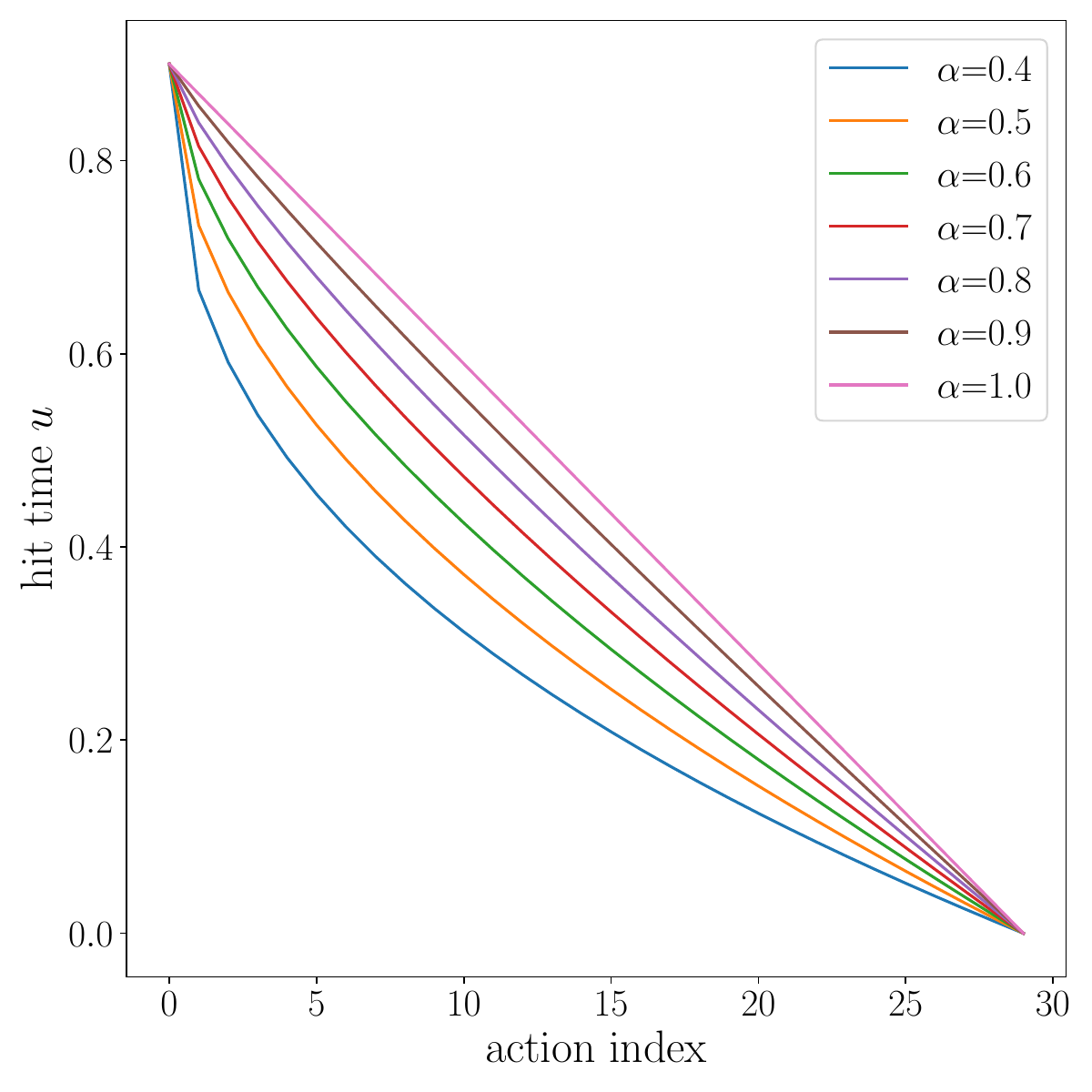}
    \vspace{-3mm}
    \caption{Hit times used in ablation study, with factor $\alpha$ from 0.4 to 1.0.}
    \label{fig:hittime}
  \end{minipage}
  \hfill
  \begin{minipage}{0.57\linewidth}
    \centering
    \captionof{table}{Ablation study of factor $\alpha$ in the Horizon-Aware Schedule, using X-VLA on the CALVIN benchmark. Mixing probability $p$ is set to 0.5.}
    \small
    \setlength{\tabcolsep}{6pt}
    \begin{tabular}{c|cccccc}
      \toprule \multirow{2}{*}{$\alpha$} & \multicolumn{6}{c}{CALVIN ABC$\rightarrow$D} \\
                                         & 1                                           & 2    & 3    & 4    & 5    & Avg. Len       \\
      \midrule 0.4                       & 96.7                                        & 91.3 & 82.5 & 73.1 & 63.5 & 4.071          \\
      0.5                                & 95.1                                        & 88.4 & 79.3 & 69.6 & 58.7 & 3.911          \\
      0.6                                & 97.5                                        & 89.5 & 80.2 & 71.2 & 60.7 & 3.991          \\
      0.7                                & 98.1                                        & 91.4 & 82.4 & 72.9 & 64.5 & \textbf{4.093} \\
      0.8                                & 95.9                                        & 88.7 & 79.7 & 71.2 & 61.5 & 3.970          \\
      0.9                                & 99.0                                        & 88.1 & 75.2 & 70.3 & 59.4 & 3.921          \\
      1.0                                & 94.7                                        & 84.0 & 71.7 & 61.3 & 51.8 & 3.635          \\
      \bottomrule
    \end{tabular}
    \label{tab:abla1}
  \end{minipage}
\end{figure}

We first investigate the factor $\alpha$ in the Horizon-Aware Schedule, which controls
the decreasing trend of hit times across the action index. As shown in
\Cref{fig:hittime}, a smaller $\alpha$ leads to a faster decay of hit times,
allocating more denoising steps to future actions. From \Cref{tab:abla1}, we
observe that HAS is robust to different values of $\alpha$, except when $\alpha =
1.0$, with the largest difference in Avg. Len being only 0.18.

\begin{table}[t]
  \centering
  \caption{Ablation study of mixing probability $p$ in the mixed schedule, using
  X-VLA on the CALVIN benchmark. Factor $\alpha$ is set to 0.7.}
  \small
  \setlength{\tabcolsep}{10pt}
  \begin{tabular}{c|cccccc}
    \toprule \multirow{2}{*}{$p$} & \multicolumn{6}{c}{CALVIN ABC$\rightarrow$D} \\
                                  & 1                                           & 2    & 3    & 4    & 5    & Avg. Len       \\
    \midrule Baseline             & 95.7                                        & 89.8 & 82.4 & 77.0 & 70.2 & 4.151          \\
    \midrule 0.3                  & 93.7                                        & 85.2 & 76.0 & 65.5 & 55.2 & 3.756          \\
    0.5                           & 98.1                                        & 91.4 & 82.4 & 72.9 & 64.5 & \textbf{4.093} \\
    0.7                           & 89.6                                        & 76.7 & 63.2 & 50.8 & 40.3 & 3.206          \\
    1.0 (w/o mixed)               & 89.0                                        & 74.7 & 60.4 & 49.0 & 38.1 & 3.112          \\
    \midrule Independent          & 91.4                                        & 82.6 & 74.0 & 64.6 & 54.5 & 3.671          \\
    \bottomrule
  \end{tabular}
  \label{tab:abla2}
\end{table}

We further analyze the influence of the mixed schedule in \method training. In
addition to varying the mixing probability $p$ between HAS and the conventional
constant schedule ($p=0$), we also include the independent time schedule proposed
in Diffusion Forcing~\cite{diffusionforcing}, where the timestep is sampled
independently across the action index during training while HAS is used for
sampling.

As shown in \Cref{tab:abla2}, a small value of $p$ leads to degraded performance,
since the inference-time schedule is observed less frequently during training. As
discussed in the main paper, fine-tuning a pretrained VLA primarily with HAS, \ie,
using a large value of $p$, can also introduce severe adverse effects,
particularly on long-horizon performance. Although training with the independent
schedule achieves reasonable accuracy, it is not directly comparable to our
training strategy: independently sampling timesteps for each action in training enlarges
the search space and can induce inconsistencies between training and inference~\cite{sun2025ar}.
These results highlight the importance of the mixed schedule in FASTER. We therefore
use the default value $p=0.5$ for all reported results and do not further tune $p$.

We also evaluate different values of $\alpha$ in real-world experiments with two
models and observe no clear performance differences. This suggests that task-specific
hyperparameter tuning is not necessary for HAS, and that our default configuration
generalizes well across different settings.

\subsection{Error Analysis}

\begin{figure}[t]
  \centering
  \includegraphics[width=0.9\linewidth]{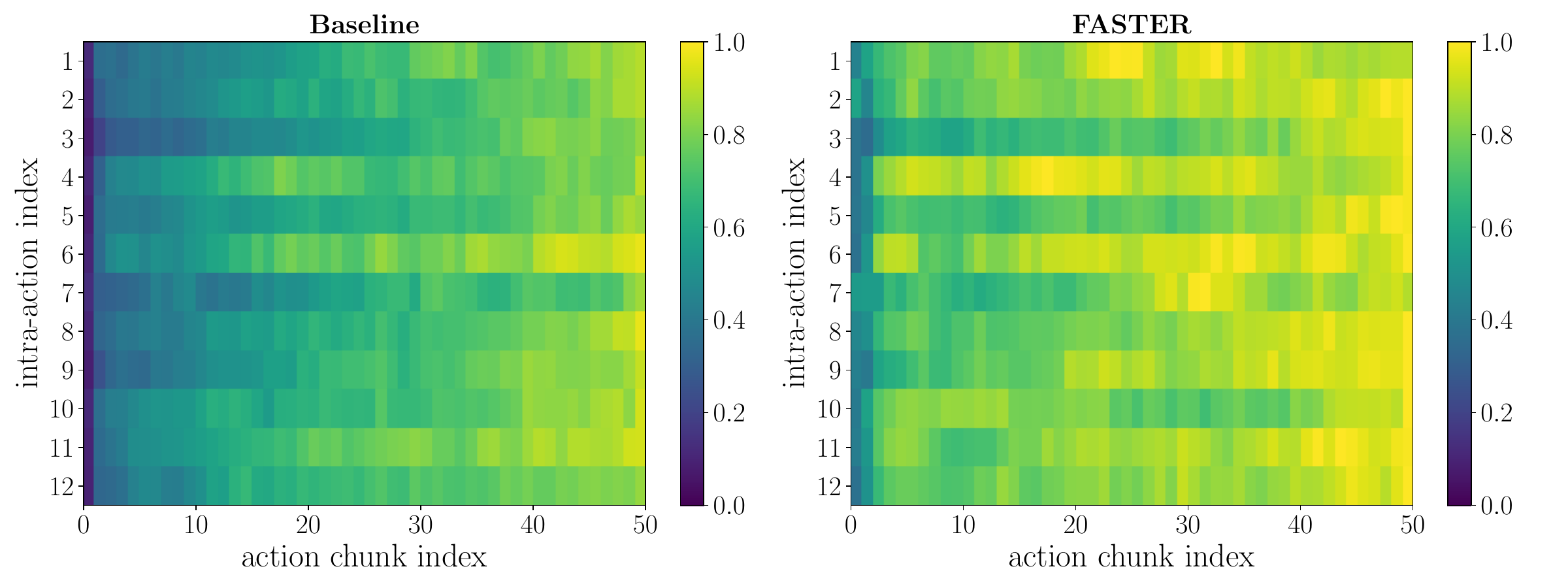}
  \caption{ Open-loop action prediction error on the Fold Towel task. We compare
  the $\pi_{0.5}$ baseline with \method. The heatmaps report the mean absolute error
  for each action dimension and action chunk index, normalized by the maximum
  error of each action dimension computed over 200 random samples and shared
  across both models.}
  \label{fig:error}
\end{figure}

We include an error analysis on real-world task to better understand how accurately
\method generates actions. We use fine-tuned $\pi_{0.5}$ models with the
constant schedule and HAS, and conduct open-loop tests to compare the inferred actions
against ground-truth demonstrations. Specifically, we measure the normalized mean
absolute error for each action dimension across different action chunk indices.

As shown in Figure~\ref{fig:error}, both methods exhibit an overall increase in
prediction error as the action index moves farther into the future, which echoes
our main insight that early actions are easier to predict. Compared with the
baseline, \method shows moderately higher errors on some action dimensions, but
these errors are distributed more uniformly across chunk indices, reflecting the
accuracy-latency trade-off introduced by accelerated sampling. Nevertheless, the
error remains relatively low for near-term actions, which are directly executed
for immediate reaction. This suggests that HAS preserves sufficient precision for
latency-critical actions while enabling substantially faster response in real-time
control.

\section{Limitations and Future Work}
\label{supp:limitation} Despite the improved reaction capability demonstrated by
FASTER, several limitations remain. First, our method is primarily applicable to
flow-based or diffusion-based VLA models whose action generation process
involves iterative sampling. The actual latency reduction also depends on the implementation
and hardware. For example, we observe that in JAX implementations, the forward-pass
runtime is not always proportional to the number of iterations, which may limit the
practical speedup obtained.

Second, FASTER is a deployment-oriented inference scheduling and streaming
framework rather than a fundamental modeling advance. Thus, it does not by
itself resolve limitations inherited from the underlying VLA model, such as perception
failures, language grounding errors, or inaccurate action generation. The robot
may still fail if the base policy produces incorrect actions for the task.

Third, our reaction time analysis relies on a simplified timing model. In
particular, we assume that sensing, communication, pre-processing, model
inference, and action dispatch can be approximated by a single effective latency
term. These assumptions lead to a uniform reaction-time distribution that
provides a useful first-order characterization of responsiveness, but may not fully
capture all deployment conditions. In practice, system latency may vary due to CPU/GPU
scheduling, network jitter, memory contention, or OS overhead.

Finally, the Horizon-Aware Schedule introduces a trade-off between responsiveness
and action accuracy. Although our experiments show that FASTER largely preserves
task performance and often improves real-world execution by reducing delay, aggressive
early sampling may slightly perturb the original generation trajectory,
especially for tasks that require precise long-horizon coordination. Better adaptive
schedules that depend on uncertainty, task phase, or online feedback may further
improve this trade-off.

\section{Broader Impacts}
\label{supp:impact} FASTER aims to improve the real-time responsiveness of VLA
models for physical robot control. Its positive impacts include enabling more reliable
robot behavior in dynamic environments, reducing the hardware requirements for
deploying generalist policies, and making real-time embodied AI systems more accessible
on consumer-grade GPUs.

At the same time, it introduces potential risks. More reactive robots can execute
actions faster in the physical world, which may amplify the severity of failures
when VLAs make incorrect predictions, misinterpret instructions, or encounter
out-of-distribution observations. The method could also contribute to broader automation
capabilities. While this can improve productivity, it also raises concerns about
labor displacement or misuse.

Our experiments are conducted in controlled research settings. We do not
recommend deploying such robot systems in safety-critical or unsupervised human-facing
environments without additional risk assessment, safety monitors, and
appropriate physical safeguards.